\RequirePackage{fix-cm}
\documentclass[twocolumn,final]{svjour3}          
\smartqed  
\usepackage[square,numbers,sort&compress]{natbib}
\usepackage[pagebackref=true,breaklinks=true,letterpaper=true,colorlinks,bookmarks=false,citecolor=blue]{hyperref}

\usepackage[utf8]{inputenc} 
\usepackage{hyperref}       
\usepackage{url}            
\usepackage{booktabs}       
\usepackage{amsfonts}       
\usepackage{nicefrac}       
\usepackage{microtype}      
\usepackage{xspace}
\usepackage{graphicx}
\usepackage{enumitem}
\usepackage{amssymb}
\usepackage{amsmath}
\usepackage{xcolor}
\usepackage[american]{babel}

\usepackage{amssymb}
\usepackage{amsmath,amsfonts}
\usepackage{amsopn,bm}

\usepackage{nicefrac}       

\usepackage{epsfig}
\usepackage{graphicx}
\usepackage{float}
\usepackage{wrapfig}
\usepackage[ruled,vlined]{algorithm2e}

\usepackage{bm,xspace}
\usepackage{comment}
\usepackage{verbatim}
\usepackage{multirow}
\usepackage{makecell}
\usepackage{array}
\usepackage{balance}
\usepackage{url}
\usepackage{booktabs}
\usepackage{etoolbox,siunitx}
\usepackage{calc}
\usepackage{pifont,hologo}
 
\usepackage{nicefrac}

\usepackage{arydshln}
\usepackage[utf8]{inputenc} 
\usepackage[T1]{fontenc}    
\usepackage{url}            
\usepackage{amsfonts}       
\usepackage{nicefrac}       
\usepackage{microtype}      
\usepackage[american]{babel}
\usepackage{enumitem}
\usepackage{siunitx}

\usepackage{enumitem}

\usepackage[super]{nth}
\usepackage{nicefrac}
\robustify\bfseries

\usepackage{dsfont}

\usepackage{listings}
\usepackage{xcolor}
\definecolor{codegreen}{rgb}{0,0.6,0}
\definecolor{codegray}{rgb}{0.5,0.5,0.5}
\definecolor{codepurple}{rgb}{0.58,0,0.82}
\definecolor{backcolour}{rgb}{1.0,1.0,1.0}

\lstdefinestyle{mystyle}{
    commentstyle=\color{codegreen},
    keywordstyle=\color{magenta},
    numberstyle=\tiny\color{codegray},
    stringstyle=\color{codepurple},
    basicstyle=\ttfamily\footnotesize,
    breakatwhitespace=false,         
    breaklines=true,                 
    captionpos=b,                    
    keepspaces=true,                 
    numbersep=0pt,                  
    showspaces=false,                
    showstringspaces=false,
    showtabs=false,                  
    tabsize=2,
    linewidth=.99\textwidth,
    xleftmargin=0.01cm
}
\usepackage{mathtools}

\newcommand{\vct}[1]{\boldsymbol{#1}} 
\newcommand{\mat}[1]{\boldsymbol{#1}} 
\newcommand{\cst}[1]{\mathsf{#1}}  

\newcommand{\field}[1]{\mathbb{#1}}
\newcommand{\R}{\field{R}} 

\newcommand{\ProbOpr}[1]{\mathbb{#1}}

\newcommand{\expect}[2]{%
\ifthenelse{\equal{#2}{}}{\ProbOpr{E}_{#1}}
{\ifthenelse{\equal{#1}{}}{\ProbOpr{E}\left[#2\right]}{\ProbOpr{E}_{#1}\left[#2\right]}}} 
\newcommand{\var}[2]{%
\ifthenelse{\equal{#2}{}}{\ProbOpr{VAR}_{#1}}
{\ifthenelse{\equal{#1}{}}{\ProbOpr{VAR}\left[#2\right]}{\ProbOpr{VAR}_{#1}\left[#2\right]}}} 

\newcommand{\vb}{\vct{b}}

\newcommand{\vp}{\vct{p}}

\newcommand{\vx}{{\vct{x}}}
\newcommand{\vy}{\vct{y}}

\newcommand{\mV}{\mat{V}}
\newcommand{\mU}{\mat{U}}

\newcommand{\cM}{\cst{M}}

\newcommand{\vphi}{\vct{\phi}}

\def\calB{\mathcal{B}}
\def\calC{\mathcal{C}}
\def\calD{\mathcal{D}}

\def\calS{\mathcal{S}}

\def\calU{\mathcal{U}}

\makeatletter
\DeclareRobustCommand\onedot{\futurelet\@let@token\@onedot}
\def\@onedot{\ifx\@let@token.\else.\null\fi\xspace}

\def\eg{\emph{e.g}\onedot} 
\def\ie{\emph{i.e}\onedot} 
\def\cf{\emph{c.f}\onedot} 
 
\def\wrt{w.r.t\onedot} 
\def\etal{\emph{et al}\onedot}
\makeatother

\newcommand{\eat}[1]{{}}

\newcommand\mypara[1]{\vspace{2mm}\noindent\textbf{#1}}

\newcommand{\castle}{\textsc{Castle}\xspace}
\newcommand{\acastle}{\textsc{a}\textsc{Castle}\xspace}

\begin{document}

\title{Learning Adaptive Classifiers Synthesis \\ for Generalized Few-Shot Learning
}

\makeatletter
\newcommand{\printfnsymbol}[1]{%
	\textsuperscript{\@fnsymbol{#1}}%
}
\makeatother


\author{Han-Jia Ye$^*$\thanks{$^*$ Equal Contribution.}  \and
        Hexiang Hu$^*$ \and
        De-Chuan Zhan
}


\institute{
        Han-Jia Ye \at
        State Key Laboratory for Novel Software Technology\\
        Nanjing University \\
        \email{yehj@lamda.nju.edu.cn}
        \and
        Hexiang Hu \at
        University of Southern California\\
        \email{hexiangh@usc.edu} 
        \and
        De-Chuan Zhan \at
        State Key Laboratory for Novel Software Technology\\
        Nanjing University \\
        \email{zhandc@lamda.nju.edu.cn}
}

\date{Received: date / Accepted: date}

\maketitle

\begin{abstract}
Object recognition in the real world requires handling long-tailed or even open-ended data. An ideal visual system needs to recognize the populated head visual concepts reliably while efficiently learning about emerging new tail categories with a few training instances.
Class-balanced many-shot learning and few-shot learning tackle one side of this problem, by either learning strong classifiers for head or learning to learn few-shot classifiers for the tail.
In this paper, we investigate the problem of \textit{generalized few-shot learning (GFSL)} ---- a model during the deployment is required to learn about tail categories with few shots and simultaneously classify the head classes.
We propose the ClAssifier SynThesis LEarning~(\castle). This learning framework learns how to synthesize calibrated few-shot classifiers in addition to the multi-class classifiers of head classes with a shared neural dictionary, shedding light upon the {\em inductive} GFSL.
Furthermore, we propose an adaptive version of \castle~(\acastle) that adapts the head classifiers conditioned on the incoming tail training examples, yielding a framework that allows effective backward knowledge transfer. As a consequence, \acastle can handle GFSL with classes from heterogeneous domains effectively.
\castle and \acastle demonstrate superior performances than existing GFSL algorithms and strong baselines on \textit{Mini}ImageNet as well as \textit{Tiered}ImageNet datasets.
More interestingly, they outperform previous state-of-the-art methods when evaluated with standard few-shot learning criteria.

\keywords{Image Recognition \and Meta Learning \and Generalized Few-Shot Learning \and Few-Shot Learning \and Recognition with Heterogeneous Visual Domain}
\end{abstract}

%
%

\section{Introduction}
\label{sec:intro}

\begin{figure*}[t]
	\centering
	\begin{tabular}{c|c}
		\includegraphics[height=0.175\textheight]{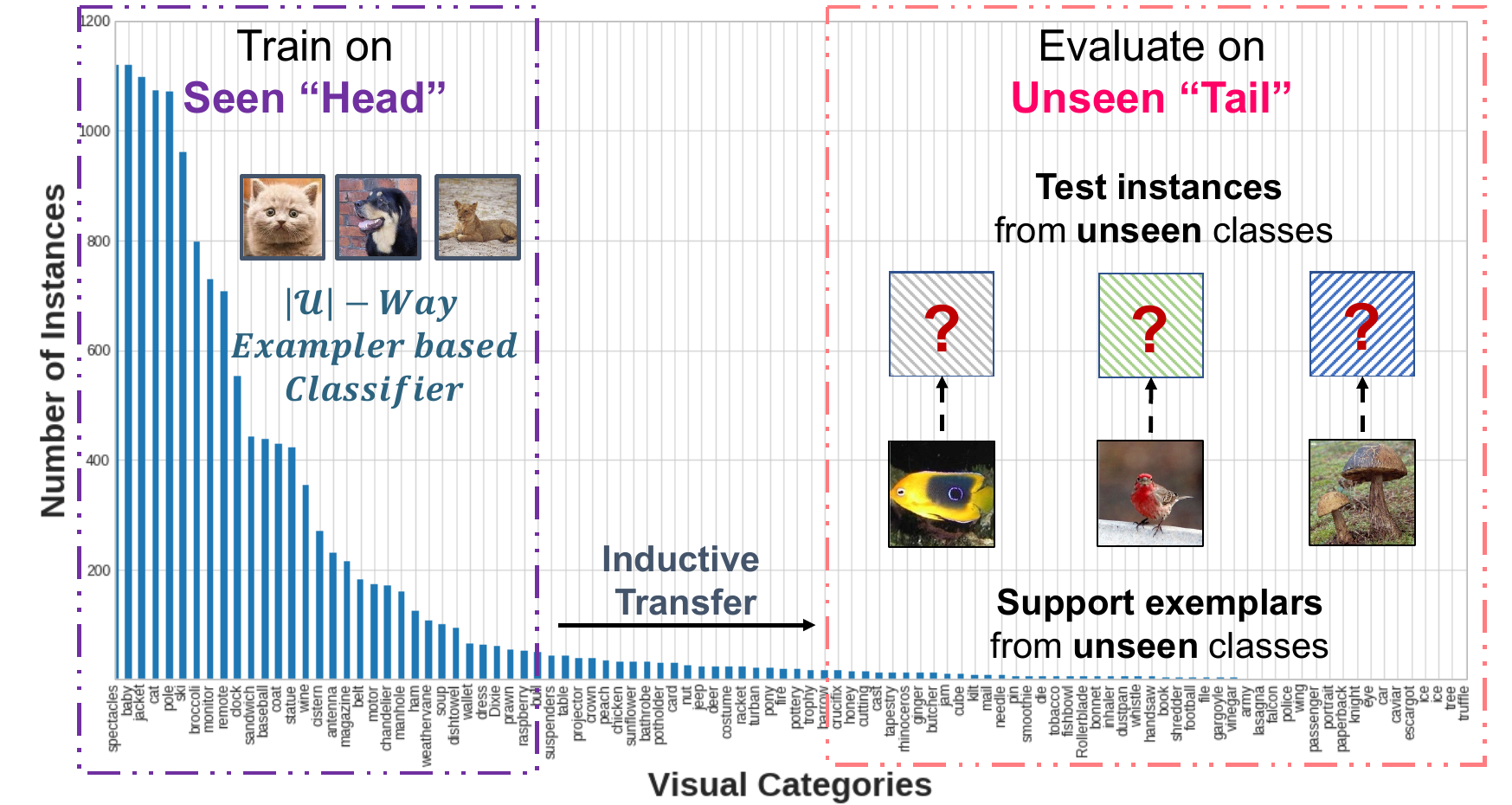} & \includegraphics[height=0.175\textheight]{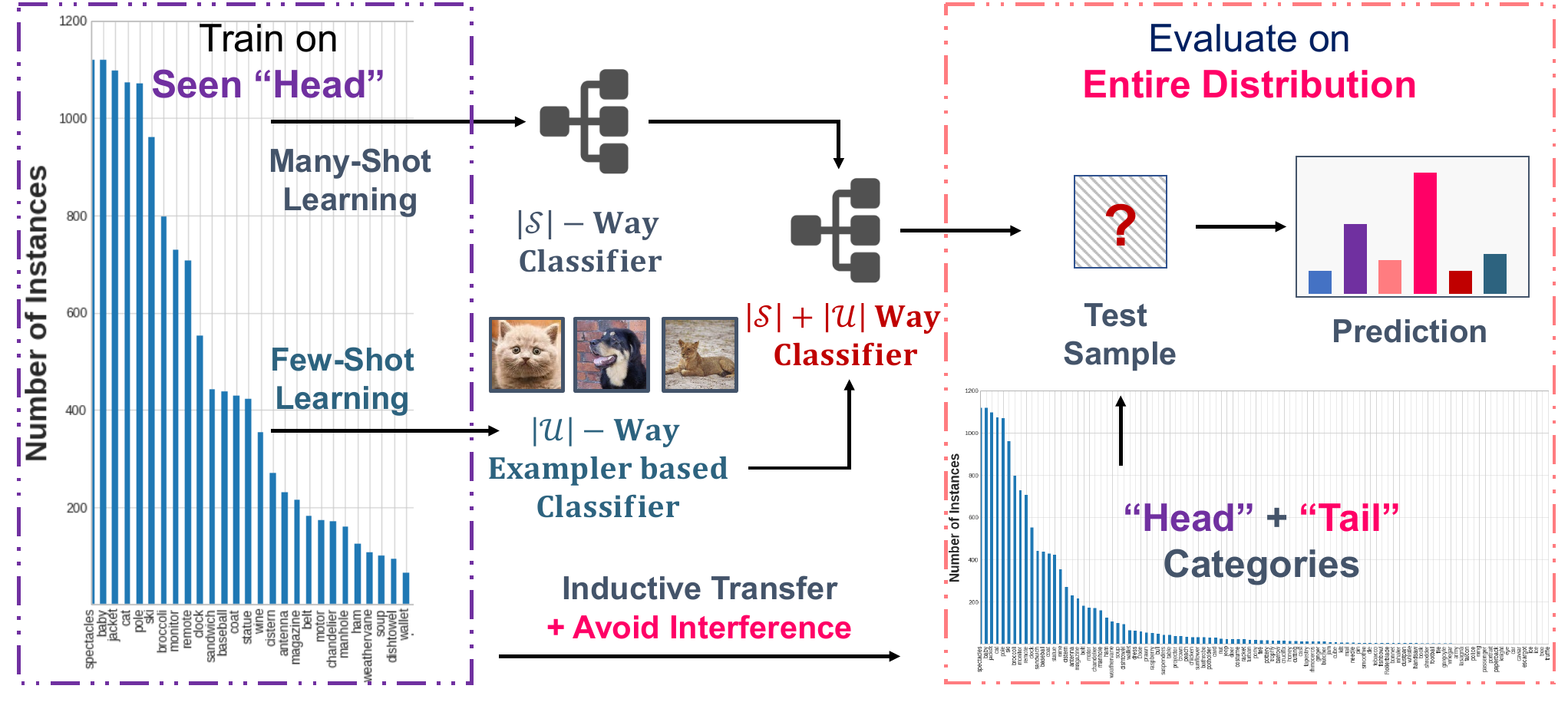} \\
		\textbf{(a)} Few-shot learning (FSL) & \textbf{(b)} Generalized Few-shot Learning (GFSL)
	\end{tabular}
	\caption{
		\textbf{A conceptual diagram comparing the Few-Shot Learning~(FSL) and the Generalized Few-Shot Learning~(GFSL).} GFSL requires to extract inductive bias from \textsc{seen} categories to facilitate efficiently learning on few-shot \textsc{unseen} tail categories, while maintaining discernability on head classes.
	}
	\label{fig:diagram}
\end{figure*}

Visual recognition for objects in the ``long tail'' has been an important challenge to address~\citep{WangRH17Learning,Liu2019Large,Kang2020Decoupling,Zhou2020BBN}. We often have a very limited amount of data on those objects as they are infrequently observed and/or visual exemplars of them are hard to collect. As such, state-of-the-art methods (\eg., deep learning) can not be directly applied due to their notorious demand of a large number of annotated data~\citep{simonyan2014very,he2016deep,Krizhevsky2017ImageNet}.

Few-shot learning~(FSL)~\citep{VinyalsBLKW16Matching,chen2019closer} is mindful of the limited data per tail concept (\ie., shots), which attempts to address this challenging problem by distinguishing between the data-rich head categories as \textsc{seen} classes and data-scarce tail categories as \textsc{unseen} classes. While it is difficult to build classifiers with data from \textsc{unseen} classes, FSL mimics the test scenarios by sampling few-shot tasks from \textsc{seen} class data, and extracts inductive biases for effective classifiers acquisition on \textsc{unseen} ones. Instance embedding~\citep{VinyalsBLKW16Matching,SnellSZ17Prototypical,Rusu2018Meta,YeHZS2018Learning}, model initialization~\citep{FinnAL17Model,Nichol2018On,Antoniou2018How}, image generator~\cite{Wang2018Low}, and optimization flow~\citep{Sachin2017,Lee2019Meta} act as popular meta-knowledge and usually incorporates with FSL.

This type of learning makes the classifier from few-shot learning for \textsc{unseen} classes difficult to be combined directly with the classifier from many-shot learning for \textsc{seen} classes, however, the demand to recognize \emph{all} object categories simultaneously in object recognition is essential as well. 

In this paper, we study the problem of \emph{Generalized Few-Shot Learning}~(GFSL), which focuses on the {\em joint} classification of both data-rich and data-poor categories. Figure~\ref{fig:diagram} illustrates the high-level idea of the GFSL, contrasting the standard FSL. 
In particular, our goal is for the model trained on the \textsc{seen} categories to be capable of incorporating the limited \textsc{unseen} class examples, and make predictions for test data in both the head and tail of the entire distribution of categories. 

One naive GFSL solution is to train a single classifier over the imbalanced long-tail distribution~\citep{Hariharan2017Low,WangRH17Learning,Liu2019Large,Zhou2020BBN}, and re-balance it~\citep{Cui2019Class,Cao2019Margin,Kang2020Decoupling}. 
One main advantage of such a joint learning objective over all classes is that it characterizes both \textsc{seen} and \textsc{unseen} classes simultaneously. In other words, training of one part (\eg., head) naturally takes the other part (\eg., tail) into consideration, and promotes the knowledge transfer between classes.
However, such a transductive learning paradigm requires collecting the limited tail data in advance, which is violated in many real-world tasks.
In contrast to it, our learning setup requires an {\em inductive} modeling of the tail, which is therefore more challenging as we assume no knowledge about the \textsc{unseen} tail categories is available during the model learning phase. 

There are two main challenges in the inductive GFSL problem, including how to construct the many-shot and few-shot classifiers in the GFSL scenario and how to \emph{calibrate} their predictions.

First, the head and tail classifiers for a GFSL model should encode different properties of all classes towards high discerning ability, and the classifiers for the many-shot part should be adapted based on the tail concepts accordingly. 
For example, if the \textsc{unseen} classes come from different domains, the same single \textsc{seen} classifier is difficult to handle their diverse properties and should not be left alone in this dynamic process.
Furthermore, as observed in the generalized zero-shot learning scenario~\cite{Chao2016Generalized}, a classifier performs over-confident with its familiar concepts and fear to make predictions for those \textsc{unseen} ones, which leads to a confidence gap when predicting \textsc{seen} and \textsc{unseen} classes. The calibration issue appears in the generalized few-shot learning as well, \ie., \textsc{seen} and \textsc{unseen} classifiers have different confidence ranges. We empirically find that directly optimizing two objectives together could not resolve the problem completely.

To this end, we propose \emph{ClAssifier SynThesis LEarning}~(\castle), where the few-shot classifiers are synthesized based on a neural dictionary with common characteristics across classes. 
Such synthesized few-shot classifiers \emph{are then used together} with the many-shot classifiers, and learned end-to-end.
To this purpose, we create a learning scenario by sampling a set of data instances from \textsc{seen} categories and pretend that they come from \textsc{unseen} categories, and apply the synthesized classifiers (based on the above instances) as if they are many-shot classifiers to optimize multi-class classification together with the remaining many-shot \textsc{seen} classifiers.
In other words, we construct few-shot classifiers to \emph{not only perform well on the few-shot classes but also to be competitive when used in conjunction with many-shot classifiers of populated classes}. 
We argue that such highly contrastive learning can benefit the few-shot classification in two aspects: (1) it provides high discernibility for its synthesized classifiers. (2) it makes the synthesized classifier automatically calibrated with the many-shot classifiers.

Taking steps further, we then propose the \emph{Adaptive ClAssifier SynThesis LEarning}~(\acastle), with additional flexibility to adapt the many-shot classifiers based on few-shot training examples. As a result, it allows backward knowledge transfer~\cite{lopez2017gradient} --- new knowledge learned from novel few-shot training examples can benefit the existing many-shot classifiers. In \acastle, the neural dictionary is the concatenation of the shared and the task-specific neural bases, whose elements summarize the generality of all visual classes and the specialty of current few-shot categories. This improved neural dictionary facilitates the adaptation of the many-shot classifiers conditioned on the limited tail training examples. 
The adapted many-shot classifiers in \acastle \emph{are then used together} with the (jointly) synthesized few-shot classifiers for GFSL classification.

We first verify the effectiveness of the synthesized GFSL classifiers over multi-domain GFSL tasks, where the \textsc{unseen} classes would come from diverse domains. {\acastle} can best handle such task heterogeneity due to its ability to adapt the head classifiers.
Next, we empirically validate our approach on two standard benchmark datasets --- {\it Mini}ImageNet~\cite{VinyalsBLKW16Matching} and {\it Tiered}ImageNet~\cite{Ren2018Meta}. 
The proposed approach retains competitive tail concept recognition performances while outperforming existing approaches on \textit{generalized} few-shot learning with criteria from different aspects. By carefully selecting a prediction bias from the validation set, those miscalibrated FSL approaches or other baselines perform well in the GFSL scenario. The implicit confidence calibration in {\castle} and {\acastle} works as well as or even better than the post-calibration techniques.
We note that {\castle} and {\acastle} are applicable for standard few-shot learning, which stays competitive with and sometimes even outperforms state-of-the-art methods when evaluated on two popular FSL benchmarks.

\mypara{Our contributions} are summarized as what follows:
\begin{itemize}[leftmargin=*]
	\item We propose a framework that synthesizes few-shot classifiers for GFSL with a shared neural dictionary, as well as its adaptive variant that modifies \textsc{seen} many-shot classifiers to allow the backward knowledge transfer.
	\item We extend an existing GFSL learning framework into an end-to-end counterpart that learns and contrasts the few-shot and the many-shot classifiers simultaneously, which is observed beneficial to the confidence calibration of these two types of classifiers.
	\item We empirically demonstrate that \acastle is effective in backward transferring knowledge when learning novel classes under the setting of multi-domain GFSL. Meanwhile, we perform a comprehensive evaluation of both existing and our approaches with criteria from various perspectives on multiple GFSL benchmarks.
\end{itemize}

In the rest sections of this paper, we first describe the problem formulation of GFSL in \S~\ref{sec:preliminary}, and then introduce our {\castle}/{\acastle} approach in \S~\ref{sec:method}. We conduct thorough experiments (see \S~\ref{sec:ext_setup} for the setups) to verify the the proposed {\castle} and {\acastle} across multiple benchmarks. We first conduct a pivot study on multi-domain GFSL benchmarks~(\S~\ref{sec:exp_cross}) to study the backward transfer capability of different methods. Then we evaluate both \acastle and \castle on popular GFSL~(\S~\ref{sec:exp_gfsl}), and FSL benchmarks~(\S~\ref{sec:exp_fsl}). Eventually, we review existing related works in \S~\ref{sec:related} and discuss the connections to our work.

\section{Problem Description}
\label{sec:preliminary}

We define a $K$-shot $N$-way task as a classification task with $N$ classes and $K$ training examples per class. The training set (\ie., the support set) is represented as $\mathcal{D}_{\mathbf{train}} = \{(\vx_{i}, \vy_{i})\}_{i=1}^{NK}$, where  $\vx_{i}\in\mathbb{R}^{D}$ is an instance and $\vy_{i}\in \{0,1\}^N$ (\ie., one-hot vector) is its label. Similarly, the test set ({\it i.e.} the query set) is $\mathcal{D}_{\mathbf{test}}$, which contains \textit{i.i.d.} samples from the distribution $\mathcal{D}_{\mathbf{train}}$.

\mypara{Many-shot learning.}
In many-shot learning where the $K$ is large (up to hundreds), a classification model $f:\mathbb{R}^D\rightarrow\{0,1\}^N$ learns by optimizing over the instances from the head classes:
\begin{equation}
\mathbb{E}_{(\vx_i,\vy_i)\in\mathcal{D}_{\mathbf{train}}} \ell(f(\vx_i), \vy_i) \nonumber
\end{equation}
Here $f$ is often instantiated as an embedding function $\vphi(\cdot):\mathbb{R}^D\rightarrow\mathbb{R}^{d}$ and a linear classifier $\mathbf{\Theta}\in\mathbb{R}^{d\times N}$: $f(\vx_i) = \vphi(\vx_i)^\top \mathbf{\Theta}$. We denote the weight vector of the $n$-class as $\mathbf{\Theta}_n$.
The loss function $\ell(\cdot, \cdot)$ measures the discrepancy between the prediction and the true label, which is typically a cross-entropy loss.

\mypara{Few-shot learning (FSL).} Different from many-shot learning, FSL faces the challenge in transferring knowledge from head visual concepts to the tail visual concepts. It assumes two non-overlapping sets of \textsc{seen} ($\mathcal{S}$) and \textsc{unseen} ($\mathcal{U}$) classes. 
The target objective is to minimize the loss over test examples of \textsc{unseen} classes:
\begin{equation}
\mathbb{E}_{\mathcal{D}^{\mathcal{U}}_{\mathbf{train}}} \mathbb{E}_{(\vx_j,\vy_j)\in\mathcal{D}^{\calU}_{\mathbf{test}}} \Big[ \ell \left(f\left(\vx_j; \mathcal{D}^{\;\calU}_{\mathbf{train}}\right), \vy_j\right) \Big]\label{eq:fsl_obj}
\end{equation}
Here, function $f$ builds the classifiers of \textsc{unseen} classes using the \textsc{unseen} training set $\mathcal{D}^{\mathcal{U}}_{\mathbf{train}}$ that minimizes loss over $\mathcal{D}^{\mathcal{U}}_{\mathbf{test}}$, denoted as $f\left(\vx_j; \mathcal{D}^{\;\calU}_{\mathbf{train}}\right)$. Given that we \textit{do not} have access to the \textsc{unseen} classes during the model training, one needs to make effective use of the \textsc{seen} classes to encode the inductive bias into the function $f$, which minimizes the objective in Eq.~\ref{eq:fsl_obj}.

\mypara{Generalized few-shot learning (GFSL).} Based on FSL, GFSL additionally aims at building a model that simultaneously predicts over $\mathcal{S} \; \cup \; \mathcal{U}$ categories. Such a model needs to deal with many-shot classification from $|\mathcal{S}|$ \textsc{seen} classes along side with learning $|\mathcal{U}|$ emerging \textsc{unseen} classes.~\footnote{$|\mathcal{S}|$ and $|\mathcal{U}|$ denote the total number of classes from the \textsc{seen} and \textsc{unseen} class sets, respectively.} The objective of GFSL is similar to the one in FSL, except that now test examples come from both \textsc{seen} and \textsc{unseen} classes:
\begin{equation}
\mathbb{E}_{\mathcal{D}^{\;\mathcal{U}}_{\mathbf{train}}} \mathbb{E}_{(\vx_j,\vy_j)\in\mathcal{D}^{\cal{S}\cup\calU}_{\mathbf{test}}} \Big[ \ell \left(f\left(\vx_j; \mathcal{D}^{\;\calU}_{\mathbf{train}}, \mathbf{\Theta}_{\calS}\right), \vy_j\right) \Big] \;\label{eq:gfsl_obj}
\end{equation}
Different from the FSL formulation (\ie, Eq.~\ref{eq:fsl_obj}), a GFSL classifier $f\left(\cdot; \mathcal{D}^{\;\calU}_{\mathbf{train}}, \mathbf{\Theta}_{\calS}\right)$ takes both the \textsc{unseen} class few-shot training set $\mathcal{D}^{\;\calU}_{\mathbf{train}}$ and the set of many-shot classifiers $\mathbf{\Theta}_{\calS}$ from the \textsc{seen} classes as input.

\subsection{Meta-Learning for Generalized Few-Shot Learning}

The goal of generalized few-shot learning is to learn the function $f$ that classifies a test instance $\vx_{j} \sim \mathcal{D}^{\calU}_{\mathbf{test}}$ as  $\hat{\vy}_{j} = f(\vx_{j}; \mathcal{D}^{\calS\;\cup\;\calU}_{\mathbf{train}})\in\{0,1\}^{|\calS|+|\mathcal{U}|}$, for the classes sampled from both $|\mathcal{S}|$ \textsc{seen} categories $\calS$ and $|\mathcal{U}|$ \textsc{unseen} categories $\calU$. During the training, a GFSL model only has access to the \textsc{seen} head classes $\mathcal{S}$, so one needs to extract knowledge of building a joint classifier over many-shot data and few-shot data using examples on the \textsc{seen} categories. 

To this purpose, we simulate many GFSL tasks from the \textsc{seen} classes. At each simulation time, we split the \textsc{seen} classes into a tail split with classes $\calC$, and treat remaining $|\calS| - |\calC|$ classes as the head split. Therefore, Eq.~\ref{eq:gfsl_obj} is transformed into:
\begin{equation}
\mathbb{E}_{\;\calD^{\;\calC}_{\mathbf{train}}\sim\;\calC}\;\mathbb{E}_{{(\vx_j, \vy_j)\in\calD^{\calS}_{\mathbf{test}}}} \ell\left(f\left(\vx_j; \mathcal{D}^{\;\calC}_{\mathbf{train}}, \mathbf{\Theta}_{\calS-\calC}\right), \vy_j\right)\;\label{eq:gfsl_meta_obj}
\end{equation}
Here, the simulated tail classes are the subset of the full \textsc{seen} classes $\calC \subset \calS$, with the number of classes to be $N$.
As a result, the function $f$ outputs a $|\calS|$-way classifier with two steps: (1) For the simulated tail split $\calC$, it follows what $f$ does in standard few-shot learning and generates the classifiers of $\calC$ using their few-shot training examples $\mathcal{D}^{\;\calC}_{\mathbf{train}}$. (2) For the head split $\calS - \calC$, this function directly makes use of the many-shot classifiers $\mathbf{\Theta}_{\calS-\calC}$ of the $\calS-\calC$ classes to generate the classifiers. Note that the loss is measured on the test examples from the entire distribution $\calD^{\calS}_{\mathbf{test}}$, which includes both head and simulated tail classes.

\section{Method}
\label{sec:method}

\begin{figure*}[t]
	\centering
	\includegraphics[width=0.975\textwidth]{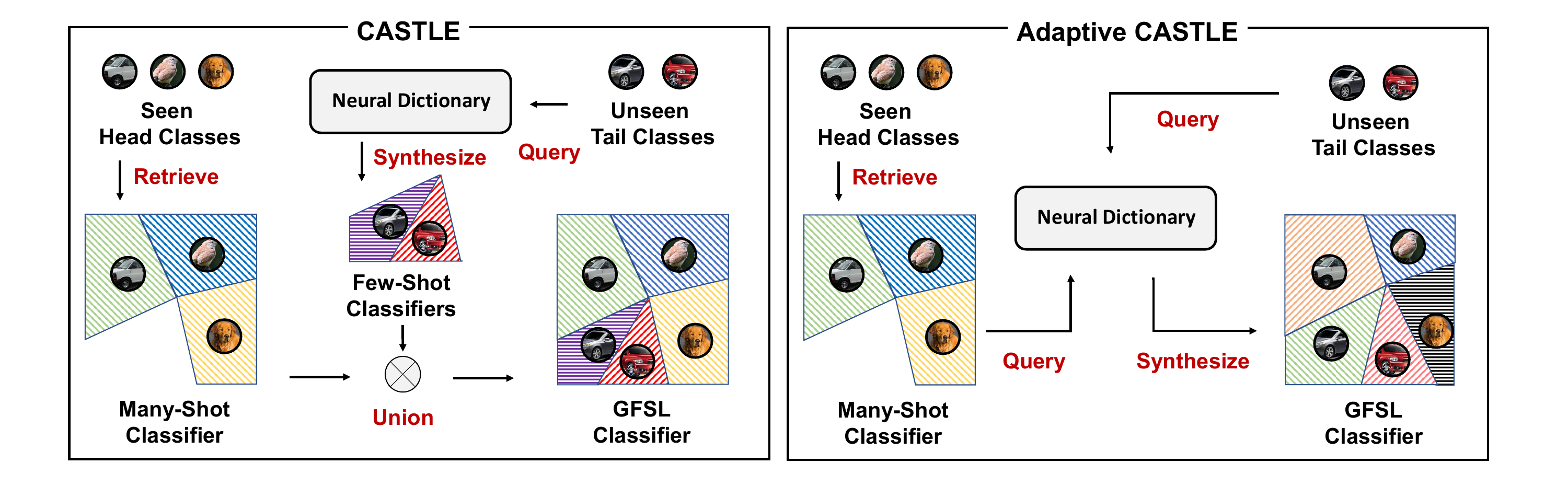}
	\caption{
	    \textbf{Illustration {\castle} \textbf{and} {\acastle}}. 
	    In \castle (left), the synthesized few-shot classifiers are directly unioned with the many-shot classifiers to make the join prediction. Different from that, {\acastle} (right) synthesizes the joint classifiers, using both the many-shot and few-shot classifiers as query to the neural dictionary. This ensures backward knowledge transfer, where many-shot head classifiers co-adapt to the few-shot tail classifiers.
	}
	\label{fig:demo}
\end{figure*}

In this section, we first present the concrete classifier composition model to generate both many-shot and few-shot classifiers, via querying a learnable neural dictionary. To simplify the understanding, we present the situation of model inferences, with the goal of composing few-shot classifiers using the 
\textsc{unseen} few-shot training data at the time of model evaluation (see \S~\ref{composition}).
Next, we introduce an effective learning algorithm that learns many-shot classifiers and few-shot classifiers simultaneously, in an end-to-end manner. We discuss the detail of how generalized few-shot learning tasks are simulated with only data of \textsc{seen} classes (see \S~\ref{ULO}).

\subsection{Classifier Composition with a Neural Dictionary}
\label{composition}

We introduce a neural dictionary based method to construct both many-shot and few-shot classifiers. This neural dictionary is being used as the common bases $\calB = \{\vb_k\}$ for constructing both head and tail classifiers. Formally, the neural bases contain two sets of elements:
\begin{equation}
    \calB = \calB_{\mathbf{share}} \bigcup \calB_{\mathbf{specific}} \nonumber
\end{equation} 
Here, $\calB_{\mathbf{share}}$ has $\cM$ learnable element $\{\vb_1, \ldots, \vb_{{\cM}}\}$ with $\vb_k \in \R^{d}$, which are globally shared for all tasks. Meanwhile, $\calB_{\mathbf{specific}}$ introduces the task-specific bases to construction of the joint classifier, as discussed later.

Based on this, we define learnable key and value embeddings, where key and value are associated with neural bases that encode shared primitives for composing the classifier of $\calS \cup \calU$. Similar to ~\cite{Vaswani2017Attention}, the key and value for the neural dictionary are generated based on two linear projections {$\mU$ and $\mV$} of elements in the bases $\calB$. For instance, $\mU \vb_k$ and $\mV \vb_k$ represent the generated key and value embeddings. For a query to the neural dictionary, it first computes the similarity (a.k.a. the attention) with all keys ($\mU \vb_k$), and the corresponding output of the query is the attention-weighted combination of all the elements in the value set ($\mV \vb_k$). 

Suppose we have sampled a $K$-shot $N$-way training data $\calD^{\;\calU}_{\mathbf{train}}$, we first compute the the class prototype of a category $c$ by taking average of all $K$ instance embeddings from the $\calD^{\calU}_{\mathbf{train}}$ (in a $K$-shot $N$-way task):
\begin{equation}
\vp_c = \frac{1}{K} \sum_{(\vx_i,\vy_i)\in\mathcal{D}^{\calU}_{\mathbf{train}}} \vphi\left(\vx_i\right) \cdot \mathbb{I}\left[\;\vy_i=c\;\right]
\label{eProto}
\end{equation}
Here, we denote $\mathbb{I}\left[\;\vy_i=c\;\right]$ as an indicator that selects instances in the class $c$. 
These class prototypes are then used as the task-specific bases $\calB_{\mathbf{specific}}$ for constructing the classifiers, which indicates that $\calB = \calB_{\mathbf{share}} \bigcup \{\vp_c \mid c \in \calU\}$.
We then compute the attention coefficients $\alpha_c$ over each bases $\vb_k \in \calB$, which are then used for assembling the classifier of class $c$.
\begin{equation}
    \alpha(\vp_c, \vb_k) \propto \exp\left(\vp_c^\top \mU \vb_k \right), \text{where \;} k = 1, \cdots, |\calB|  \nonumber
\end{equation}
The coefficient $\alpha^k_c$ is then \textit{normalized} with the sum of compatibility scores over all $|\calB|$ bases, which then is used to convexly combine the value embeddings and synthesize the classifier,
\begin{equation}
\mathbf{W}_c = \vp_c + \sum_{k=1}^{|\calB|} \alpha(\vp_c, \vb_k) \cdot \mV \vb_k\label{eq:sythesize}
\end{equation}
As a result, $\mathbf{W}_\calU=\{\mathbf{W}_c\}_{c\in\calU}$ represents the synthesized classifiers for the few-shot task using $\calD^{\calU}_{\mathbf{train}}$. Such a composed classifier is then $\ell_2$-normalized and unions together with the many-shot classifiers $\mathbf{\Theta}_{\calS}$ to make joint prediction over all classes $\calS \cup \calU$:
\begin{equation}
    \hat{\mathbf{W}} = \mathbf{\Theta}_{\calS} \cup \mathbf{W}_\calU
\end{equation}
We denote the above classifier composition method as the ClAssifier SynThesis LEarning (\castle), and illustrate its conceptual inference procedure in the left part of the Figure~\ref{fig:demo}. 

\mypara{The adaptive \castle (\acastle).} One important limitation of \castle is its lack of adaptation in the many-shot classifiers $\mathbf{\Theta}_{\calS}$. To overcome this drawback, we further propose an adaptive version of \castle, denoted as \acastle, where we use the neural dictionary to synthesize classifiers for both many-shot \textsc{seen} categories $\calS$ and few-shot \textsc{unseen} categories $\calU$. The right part of the Figure~\ref{fig:demo} illustrates the conceptual inference procedure of \acastle.

Comparing to \castle, \acastle presents two key differences. First, it additionally includes the many-shot classifiers to the task-specific bases of the neural dictionary:
\begin{equation}
    \calB_{\mathbf{specific}} = \{\vp_c \mid c \in \calU\} \bigcup \mathbf{\Theta}_{\calS}
\end{equation}
which means that the synthesis of few-shot classifier can better leverage the context of many-shot classifiers. More importantly, all the classifiers $\mathbf{W}_c,\; \forall c \in \calS \cup \calU$ are synthesized using the dictionary:
\begin{align}
    \hat{\mathbf{W}}_c = \Bigg\{
    \begin{array}{lr}
        \vp_c + \sum_{k=1}^{|\calB|} \alpha(\vp_c, \vb_k) \cdot \mV \vb_k, & \forall c \in \calU \\[8pt]
        \mathbf{\Theta}_c + \sum_{k=1}^{|\calB|} \alpha(\mathbf{\Theta}_c, \vb_k) \cdot \mV \vb_k, & \forall c \in \calS 
    \end{array}    
\end{align}
Here, $\hat{\mathbf{W}}$ corresponds to the final joint classifier. 
We specially note that \acastle allows backward knowledge transfer, from the few-shot training data $\calD^{\calU}_{\textbf{train}}$ to the many-shot classifiers $\mathbf{\Theta}_{\calS}$. We note that \castle can be regarded as a degenerated version of the \acastle.

\subsection{Unified Learning of Few-Shot and Many-Shot Classifiers}
\label{ULO}

In the GFSL, few-shot classifiers are required to classify images together with many-shot classifiers. Suppose we have sampled a $K$-shot $N$-way few-shot learning task $\calD^{\;\calU}_{\mathbf{train}}$, which contains $|\calU|$ visual \textsc{unseen} categories, a GFSL classifier $f$ should have a low expected error as in Eq.~\ref{eq:gfsl_obj}. 
Specifically, as aforementioned in \S~\ref{composition}, we use the neural dictionary to implement the joint classifier $f\left(\vx_j; \mathcal{D}^{\;\calU}_{\mathbf{train}}, \mathbf{\Theta}_{\calS}\right)$ via either \castle or \acastle.
Then the classifier $f$ predicts a test example in $\calD^{\;\calS\;\cup\;\calU}_{\mathbf{test}}$ as \emph{both} tail classes $\calU$ and head classes $\calS$. However, since we have no access to the \textsc{unseen} classes $\calU$, the learning can only happen on the data from \textsc{seen} classes $\calS$. Similar to the meta-learning procedure of standard FSL, we can simulate the learning situation of generalized few-shot learning using the data from the \textsc{seen} categories $\calS$, to mimic both head and tail classes. 

\mypara{Unified learning objective.} 
Suppose we sample a $K$-shot $N$-way few-shot task with categories $\calC$ to simulate the GFSL learning situation, where $\calC$ is a subset of \textsc{seen} classes $\calS$. Therefore, given the simulated few-shot task, we treat the remaining $\calS-\calC$ classes as the simulated head classes, whose corresponding many-shot classifiers are ${\mathbf{\Theta}}_{\calS - \calC}$. With either \castle or \acastle, we can then obtain the join classifiers $\hat{\mathbf{W}}$, that is used for predicting examples from all $\calS$ classes.
As a result, we optimize the learning objective as follows:

\begin{equation}
\min_{\{\vphi, \calB, \{\mathbf{\Theta}_s\}, \mU, \mV\}}\; \sum_{\calC \subset \calS} \; {{\sum}_{{(\vx_j, \vy_j)\sim\mathcal{S}}}} \ell\Big(\hat{\mathbf{W}}^\top \vphi\big(\vx_j\big), \vy_j\Big)\;\label{eq:gfsl_final}
\end{equation}
In addition to the learnable neural bases $\calB$, $\mU$ and $\mV$ are two projections in the neural bases to facilitate the synthesis of the classifier, and there is no bias term in our implementation.
Despite that the few-shot classifiers $\hat{\mathbf{W}}_\calC$ are synthesized using with $K$ training instances, they are optimized to perform well on all the instances from $\calC$ and moreover, to perform well against all the instances from other \textsc{seen} categories $\calS - \calC$. 

\mypara{Reusing many-shot classifiers.} We optimize Eq.~\ref{eq:gfsl_final} by using the many-shot classifier over $\calS$ to initialize the embedding $\vphi$. In detail, a $|\calS|$-way many-shot classifier is trained over all \textsc{seen} classes with the cross-entropy loss, whose backbone is used to initialize the embedding $\vphi$ in the GFSL classifier. We empirically observed that such initialization is essential for the prediction calibration between \textsc{seen} and \textsc{unseen} classes.

\mypara{Multi-classifier learning.} A natural way to minimize Eq.~\ref{eq:gfsl_final} implements a stochastic gradient descent step in each mini-batch by sampling one GFSL task, which contains a $K$-shot $N$-way training set together with a set of test instances $(\vx_j, \vy_j)$ from $\cal{S}$. It is clear that increasing the number of GFSL tasks per gradient step can improve the optimization stability. With this observation, we propose an efficient implementation to utilize \emph{a large number of} GFSL tasks for gradient computation. 
Specifically, we sample two sets of instances from \emph{all} \textsc{seen} classes, \ie., $\mathcal{D}^\calS_{\mathbf{train}}$ and $\mathcal{D}^\calS_{\mathbf{test}}$. Then we construct a large number of joint classifiers $\hat{\mathbf{W}}^z$ with either \castle or \acastle on different sets of $\calC$, which is then applied to compute the averaged loss over $z$ using Eq.~\ref{eq:gfsl_final}.
Only one single model forward is required to get the instance embeddings. We mimic multiple GFSL tasks through different random partitions of the simulated few-shot and many-shot classes.

\section{Experimental Setups}
\label{sec:ext_setup}
This section details the experimental setups,  including the general data splits strategy, the pre-training technique, the specifications of the feature backbone, and the evaluation metrics for GFSL.

\begin{figure*}[tbp]
	\begin{center}
		\includegraphics[width=1.0\textwidth]{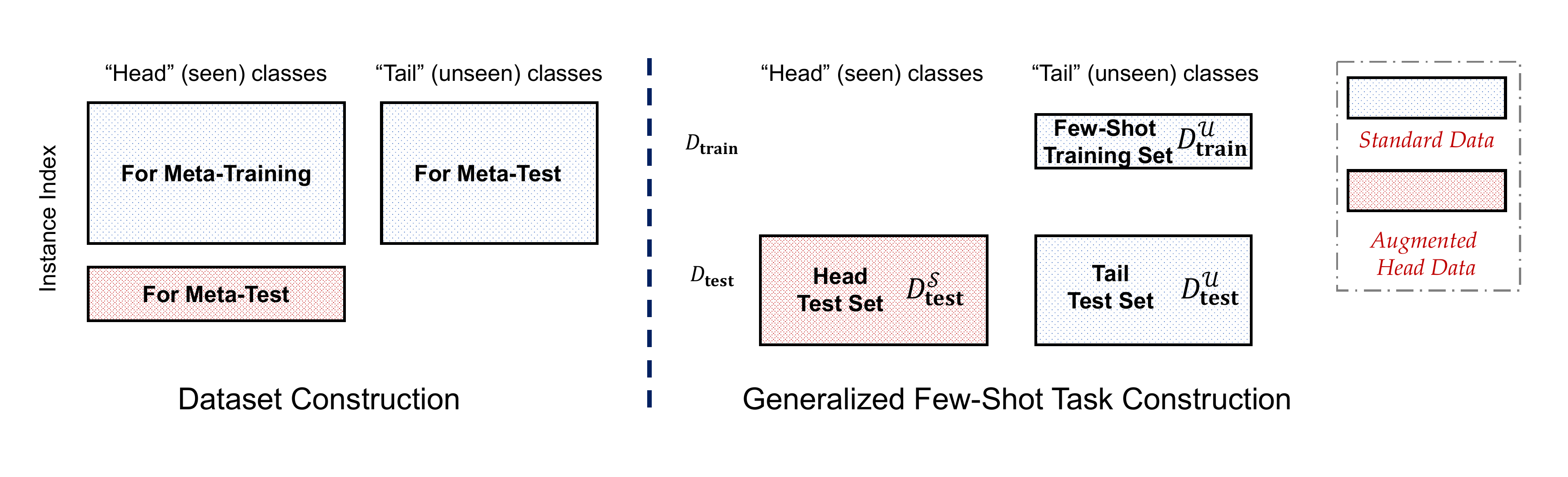}
		\caption{The split of data in the generalized few-shot classification scenario. In addition to the standard dataset like {\it Mini}Imagetnet (blue part), we collect non-overlapping augmented head class instances from the corresponding categories in the ImageNet (red part), to measure the classification ability on the \textsc{seen} classes. Then in the generalized few-shot classification task, few-shot instances are sampled from each of the \textsc{unseen} classes, while the model should have the ability to predict instances from {\em both} the head and tail classes.}
		\label{fig:data_split}
	\end{center}
\end{figure*}

\mypara{Data splits.}
We visualize the general data split strategy in Figure~\ref{fig:data_split}. There are two parts of the dataset for standard meta-learning tasks. The meta-training set for model learning (corresponds to the \textsc{seen} classes), and the meta-val/test part for model evaluation (corresponds to the \textsc{unseen} classes). To evaluate a GFSL model, we'd like to augment the meta-training set with new instances, so that the classification performance on \textsc{seen} classes could be measured. During the inference phase, a few-shot training set from \textsc{unseen} classes are provided with the model, and the model should make a joint prediction over instances from {\em both} the head and tail classes.
We will describe the detailed splits for particular datasets in later sections.

\mypara{Pre-training strategy.}
Before the meta-training stage, we try to find a good initialization for the embedding $\vphi$, and then we reuse such a many-shot classifier as well as the embedding to facilitate the training of a GFSL model. In later sections, we will verify this pre-training strategy does not influence the few-shot classification performance a lot, but it is essential to make the GFSL classifier well-calibrated. In particular, on {\it Mini}ImageNet, we add a linear layer on the backbone output and optimize a 64-way classification problem on the meta-training set with the cross-entropy loss function. Stochastic gradient descent with an initial learning rate 0.1 and momentum 0.9 is used to complete such optimization. The 16 classes in {\it Mini}ImageNet for model selection also assist the choice of the pre-trained model. After each epoch, we use the current embedding and measures the nearest neighbor based few-shot classification performance on the sampled few-shot tasks from these 16 classes. The most suitable embedding function is recorded. After that, such a learned backbone is used to initialize the embedding part $\vphi$ of the whole model. The same strategy is also applied to the meta-training set of the {\it Tiered}ImageNet, Heterogeneous, and Office-Home datasets, where 351-way, 100-way, and 25-way classifiers are pre-trained.

\mypara{Feature network specification.}
Following the setting of most recent methods~\citep{Qiao2017Few, Rusu2018Meta, YeHZS2018Learning}, we use ResNet variants~\citep{he2016deep,Bertinetto2019Meta} to implement the embedding backbone $\vphi$. We follow~\citep{Qiao2017Few,Rusu2018Meta} when investigating the multi-domain GFSL, where images are resized to $84\times84\times3$. In concrete words, three residual blocks are used after an initial convolutional layer (with stride $1$ and padding $1$) over the image, which have channels $160/320/640$, stride $2$, and padding $2$. After a global average pooling layer, it leads to a $640$ dimensional embedding. While for the benchmark experiments on {\it Mini}ImageNet and {\it Tiered}ImageNet, we follow~\cite{Lee2019Meta} to set the architecture of ResNet, which contains 12 layers and uses the DropBlock~\citep{Ghiasi2018Dropblock} to prevent over-fitting.

We use the pre-trained backbone to initialize the embedding part $\vphi$ of a model for {\castle}/{\acastle} as well as our re-implemented comparison methods such as MC+$k$NN, ProtoNet+ProtoNet, MC+ProtoNet, L2ML~\citep{WangRH17Learning}, and DFSL~\citep{Gidaris2018Dynamic}. When there exists a backbone initialization, we set the initial learning rate as 1e-4 and optimize the model with Momentum SGD. The learning rate will be halved after optimizing 2,000 mini-batches. During meta-learning, all methods are optimized over 5-way few-shot tasks, where the number of shots in a task is consistent with the inference (meta-test) stage. For example, if the goal is a 1-shot 5-way model, we sample 1-shot 5-way $\calD^\calS_\mathbf{train}$ during meta-training, together with 15 instances per class in $\calD^\calS_\mathbf{test}$.

For {\castle}/{\acastle}, we use a multi-classifier training technique to improve learning efficiency. Specifically, we randomly sample a 24-way task from $\calS$ in each mini-batch, and re-sample 64 5-way tasks from it. It is notable that all instances in the 24-way task are encoded by the ResNet backbone with the same parameters in advance. Therefore, by embedding the synthesized 5-way few-shot classifiers into the global many-shot classifier, it results in 64 different configurations of the generalized few-shot classifiers. To evaluate the classifier, we randomly sample instances with batch size 128 from $\calS$ and compute the GFSL objective in Eq.~\ref{eq:gfsl_meta_obj}.

\subsection{Evaluation Measures}
We take advantage of the auxiliary meta-training set from the benchmark datasets during GFSL evaluations, and an illustration of the dataset construction can be found in Figure~\ref{fig:data_split}. The notation $X\rightarrow Y$ with $X,Y\in\{\calS, \calU, \calS\cup\calU\}$ means computing prediction results with instances from $X$ to labels of $Y$. For example, $\calS\rightarrow\calS\cup\calU$ means we first filter instances come from the \textsc{seen} class set ($\vx\in\calS$), and predict them into the joint label space ($\vy\in\calS\cup\calU$).
For a GFSL model, we consider its performance with different measurements. 

\mypara{Few-shot accuracy.} Following the standard protocol~\citep{VinyalsBLKW16Matching,FinnAL17Model,SnellSZ17Prototypical,YeHZS2018Learning}, we sample 10,000 $K$-shot $N$-way tasks from $\calU$ during inference. In detail, we first sample $N$ classes from $\calU$, and then sample $K+15$ instances for each class. The first $NK$ labeled instances ($K$ instances from each of the $N$ classes) are used to build the few-shot classifier, and the remaining $15N$ (15 instances from each of the $N$ classes) are used to evaluate the quality of such few-shot classifier. During our test, we consider $K=1$ and $K=5$ as in the literature, and change $N$ ranges from $\{5,10,15,\ldots,|\calU|\}$ as a more robust measure. It is noteworthy that in this test stage, all the instances come from $\calU$ and are predicted to classes in $\calU$ ($\calU \rightarrow \calU$).

\mypara{Generalized few-shot accuracy.} Different from many-shot and few-shot evaluations, the generalized few-shot learning takes the joint instance and label spaces into consideration. In other words, the instances come from $\calS\cup\calU$ and their predicted labels also in $\calS\cup\calU$ ($\calS\cup\calU \rightarrow \calS\cup\calU$). This is obviously more difficult than the many-shot~($\calS \rightarrow \calS$) and few-shot~($\calU \rightarrow \calU$) tasks. During the test, with a bit abuse of notations, we sample $K$-shot $\mathcal{S}+N$-way tasks from $\calS \cup \calU$. Concretely, we first sample a $K$-shot $N$-way task from $\calU$, with $NK$ training and $15N$ test instances, respectively. Then, we {\em randomly} sample $15N$ instances from $\calS$. Thus in a GFSL evaluation task, there are $NK$ labeled instances from $\calU$, and $30N$ test instances from $\calS \cup \calU$. We compute the accuracy of $\calS\cup\calU$ as the final measure. We abbreviate this criterion as ``{\it Mean Acc.}'' or ``{\it Acc.}''.

\begin{figure*}[tbp]
	\begin{center}
		\includegraphics[width=1.0\textwidth]{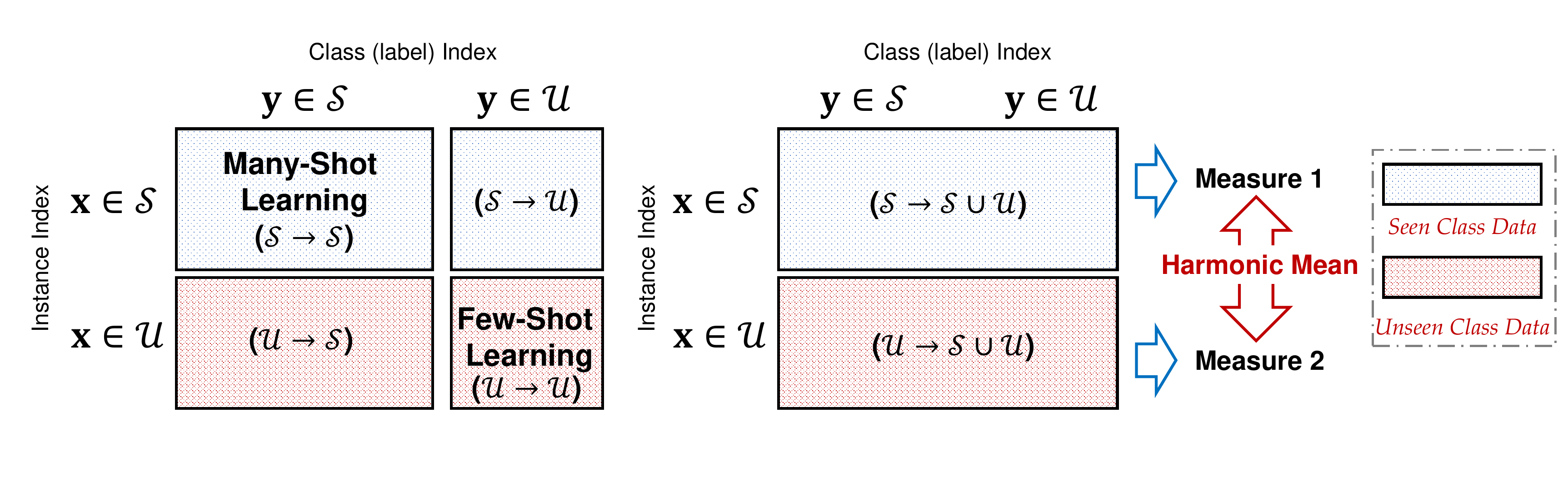}
		\caption{An illustration of the harmonic mean based criterion for GFSL evaluation. $\calS$ and $\calU$ denotes the \textsc{seen} and \textsc{unseen} instances ($\vx$) and labels ($\vy$) respectively. $\calS\cup\calU$ is the joint set of $\calS$ and $\calU$. The notation $X\rightarrow Y, X,Y\in\{\calS, \calU, \calS\cup\calU\}$ means computing prediction results with instances from $X$ to labels of $Y$.        
		By computing a performance measure (like accuracy) on the joint label space prediction of \textsc{seen} and \textsc{unseen} instances separately, a harmonic mean is computed to obtain the final measure.}
		\label{fig:hmean}
	\end{center}
\end{figure*}

\mypara{Generalized few-shot $\Delta$-value.} Since the problem becomes difficult when the predicted label space expands from $\calS \rightarrow \calS$ to $\calS \rightarrow \calS\cup\calU$ (and also $\calU \rightarrow \calU$ to $\calU \rightarrow \calS\cup\calU$), the accuracy of a model will have a drop. To measure how the classification ability of a GFSL model changes when working in a GFSL scenario, Ren \etal.~\cite{Ren2019Incremental} propose the $\Delta$-Value to measure the average accuracy drop. In detail, for each sampled GFSL task, we first compute its many-shot accuracy ($\calS \rightarrow \calS$) and few-shot accuracy ($\calU \rightarrow \calU$). Then we calculate the corresponding accuracy of \textsc{seen} and \textsc{unseen} instances in the joint label space, \ie., $\calS \rightarrow \calS\cup\calU$ and  $\calU \rightarrow \calS\cup\calU$. The $\Delta$-Value is the average decrease of accuracy in these two cases. We denote this measure as ``$\Delta$-value''. 

\mypara{Generalized few-shot harmonic mean.} Directly computing the accuracy still gets biased towards the populated classes, so we also consider the harmonic mean as a more balanced measure~\citep{Xian2017Zero}. By computing performance measurement such as top-1 accuracy for $\calS \rightarrow \calS\cup\calU$ and $\calU \rightarrow \calS\cup\calU$, the harmonic mean is used to average the performance in these two cases as the final measure. In other words, denote the accuracy for $\calS \rightarrow \calS\cup\calU$ and $\calU \rightarrow \calS\cup\calU$ as Acc$_\calS$ and Acc$_\calU$, respectively, the value $\frac{2{\rm Acc}_\calS{\rm Acc}_\calU}{{\rm Acc}_\calS + {\rm Acc}_\calU}$ is used as a final measure. An illustration is in Figure~\ref{fig:hmean}. We denote this measure as ``{\it HM}'' or ``{\it HM Acc.}''. 

\mypara{Generalized few-shot AUSUC.} Chao \etal.~\cite{Chao2016Generalized} propose a calibration-agnostic criterion for generalized zero-shot learning. To avoid evaluating a model influenced by a calibration factor between \textsc{seen} and \textsc{unseen} classes, they propose to determine the range of the calibration factor for all instances at first, and then plot the \textsc{seen}-\textsc{unseen} accuracy curve based on different configurations of the calibration values. Finally, the area under the \textsc{seen}-\textsc{unseen} curve is used as a more robust criterion. We follow~\citep{Chao2016Generalized} to compute the AUSUC value for sampled GFSL tasks. We denote this measure as ``AUSUC''.

\section{Pivot Study on Multi-Domain GFSL}
\label{sec:exp_cross}
\begin{figure*}[tbp]
	\begin{center}
		\tabcolsep 2pt
		\includegraphics[width=0.9\textwidth]{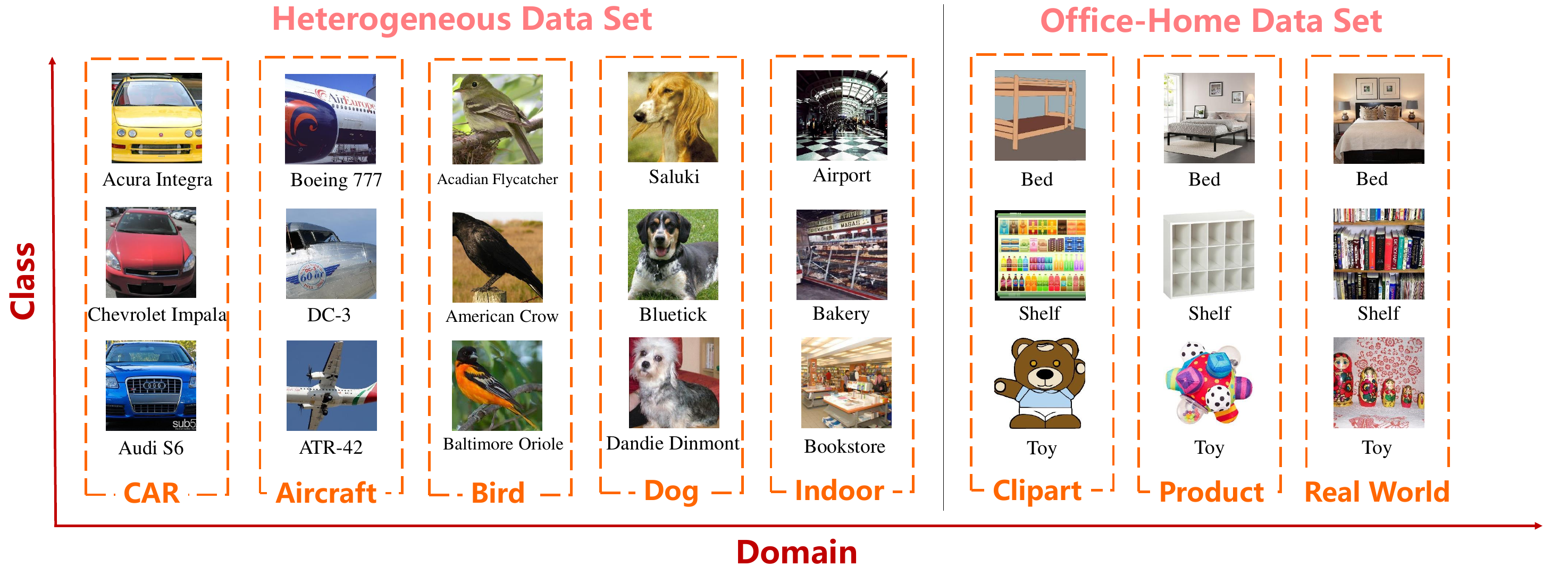}
		\caption{An illustration of the Heterogeneous and Office-Home dataset. Both datasets contain multiple domains. In the Heterogeneous dataset, each class belongs to only one domain, while in Office-Home, a class has images from all three domains.}
		\label{fig:data_set}
	\end{center}
\end{figure*}

We first present a pivot study to demonstrate the effectiveness of {\acastle}, which leverages adaptive classifiers synthesized for both \textsc{seen} and \textsc{unseen} classes. To achieve this, we investigate two multi-domain datasets -- ``Heterogeneous'' and ``Office-Home'' with more challenging settings, where a GFSL model is required to \textit{transfer knowledge in backward direction} (adapt \textsc{seen} classifiers based on \textsc{unseen} ones) to obtain superior joint classification performances over heterogeneous domains.

\begin{table*}[t]
	\centering
	\tabcolsep 2pt
 	\caption{
		Generalized 1-shot classification performance (mean accuracy and harmonic mean accuracy) on (a) the Heterogeneous dataset with {\bf 100 Head and 5 Tail categories} and (b) the Office-Home dataset with {\bf 25 Head and 5 Tail categories}. 
		$\calS\rightarrow \calS\cup\calU$ and $\calU\rightarrow \calS\cup\calU$ denote the joint classification accuracy for \textsc{seen} class and \textsc{unseen} class instances respectively. 
		\textsc{Castle}$^-$ is the variant of {\castle} without using the neural dictionary.
	}
	\begin{tabular}{cc}
	     \bf (a) Heterogeneous dataset & \bf (b) Office-Home dataset \\
	     \begin{minipage}[t]{0.485\linewidth}
	     \resizebox{\linewidth}{!}
	     {
	        \centering
	     	\tabcolsep 2pt
        	\begin{tabular}{@{\;}l@{\;}cccc@{\;}}
        		\addlinespace
        		\toprule
        		Measures & {\it $\calS\cup\calU \rightarrow \calS\cup\calU$}  & {\it $\calS\rightarrow \calS\cup\calU$} & {\it $\calU\rightarrow \calS\cup\calU$} & {\it HM Acc.} \\ \midrule
        		DFSL~\cite{Gidaris2018Dynamic} & 48.13{\tiny $\pm$ 0.12} & 46.33{\tiny $\pm$ 0.12} & 48.25{\tiny $\pm$ 0.22} & 47.27{\tiny $\pm$ 0.12}\\
        		\textsc{Castle}$^-$ & 48.29{\tiny $\pm$ 0.12} & 45.13{\tiny $\pm$ 0.13}  & 50.14{\tiny $\pm$ 0.22} & 47.50{\tiny $\pm$ 0.12} \\		
        		{\castle} & 50.16{\tiny $\pm$ 0.13}  & 48.05{\tiny $\pm$ 0.13} & \bf 50.86{\tiny $\pm$ 0.22} & 49.05{\tiny $\pm$ 0.12} \\
        		{\acastle} & \bf 53.01{\tiny $\pm$ 0.12} & \bf 56.18{\tiny $\pm$ 0.12} & 49.84{\tiny $\pm$ 0.22} & \bf 52.81{\tiny $\pm$ 0.13} \\
        		\bottomrule
        	\end{tabular}
        	\label{tab:hetero_gfsl}
        }
	    \end{minipage}
	    & \begin{minipage}[t]{0.485\linewidth}
	    \resizebox{\linewidth}{!}
	    {
	        \centering
	     	\tabcolsep 2pt
            \begin{tabular}{@{\;}l@{\;}cccc@{\;}}
                \addlinespace
            	\toprule
            	{Measures} & {\it $\calS\cup\calU \rightarrow \calS\cup\calU$}  & {\it $\calS\rightarrow \calS\cup\calU$} & {\it $\calU\rightarrow \calS\cup\calU$} & {\it HM Acc.} \\
            	\midrule
            	DFSL~\cite{Gidaris2018Dynamic} & 35.72{\tiny $\pm$ 0.12} & 28.42{\tiny $\pm$ 0.12} & 39.77{\tiny $\pm$ 0.22} & 33.15{\tiny $\pm$ 0.12}\\
            	\textsc{Castle}$^-$ & 35.74{\tiny $\pm$ 0.13} & 27.93{\tiny $\pm$ 0.13} & \bf 42.59{\tiny $\pm$ 0.22} & 33.73{\tiny $\pm$ 0.13} \\	
            	{\castle} & 35.77{\tiny $\pm$ 0.13} & 29.03{\tiny $\pm$ 0.13} & 42.46{\tiny $\pm$ 0.22} & 34.48{\tiny $\pm$ 0.13} \\
            	{\acastle} & \bf 39.99{\tiny $\pm$ 0.14} & \bf 40.29{\tiny $\pm$ 0.13} & 39.68{\tiny $\pm$ 0.22} & \bf 39.98{\tiny $\pm$ 0.14} \\
            	\bottomrule
            \end{tabular}
            \label{tab:office_gfsl}
        }
	    \end{minipage}
	\end{tabular}
\end{table*}

\subsection{Dataset}
We construct a \textbf{Heterogeneous} dataset based on 5 fine-grained classification datasets, namely AirCraft~\cite{maji13finegrained}, Car-196~\cite{CAR_data},  Caltech-UCSD Birds (CUB) 200-2011~\cite{WahCUB_200_2011}, Stanford Dog~\cite{Dog_data}, and Indoor Scenes~\cite{Quattoni2009Recognizing}. 
Since these datasets have apparent heterogeneous semantics, we treat images from different datasets as different domains. 20 classes with 50 images in each of them are randomly sampled from each of the 5 datasets to construct the meta-training set. The same sampling strategy is also used to sample classes for model validation (meta-val) and evaluation (meta-test) sets. Therefore, there are 100 classes for meta-training/val/test sets, which contain 20 classes from each fine-grained dataset. To evaluate the performance of a GFSL model, we augment the meta-training set by sampling another 15 images from the corresponding classes for each of the \textsc{seen} classes.

We also investigate the \textbf{Office-Home}~\cite{Venkateswara2017Office} dataset, which originates from a domain adaptation task. There are 65 classes and 4 domains of images per class. Considering the scarce number of images in one particular domain, we select three of the four domains, ``Clipart'', ``Product'', and ``Real World'' to construct our dataset. The number of instances in a class per domain is not equal. We randomly sample 25 classes (with all selected domains) for meta-training, 15 classes for meta-validation, and the remaining 25 classes are used for meta-test. Similarly, we hold out 10 images per domain for each \textsc{seen} class to evaluate the generalized classification ability of a GFSL model.

Note that in addition to the class label, images in these two datasets are also equipped with {\it at least one} domain label. In particular, classes in Heterogeneous dataset belong to a single domain corresponding to ``aircraft'', ``bird'', ``car'', ``dog'', or ``indoor scene'', while the classes in Office-Home possess images from all 3 domains, namely ``Clipart'', ``Product'' and ``Real World''. An illustration of the sampled images (of different domains) from these two datasets is shown in Figure~\ref{fig:data_set}.

\par\noindent\textbf{The key difference to standard GFSL} (\cf \S~\ref{sec:exp_gfsl}) is that here the \textsc{seen} categories are collected from multiple (heterogeneous) visual domains and used for training the inductive GFSL model. During the evaluation, the few-shot training instances of tail classes \textit{only come from one single domain}. With this key difference, we note that the \textsc{unseen} few-shot classes are close to a certain sub-domain of \textsc{seen} classes and relatively far away from the others. Therefore, a model capable of adapting its \textsc{seen} classifiers can take the advantages and adapt itself to the domain of the \textsc{unseen} classes.

\subsection{Baselines and Comparison Methods}
Besides {\castle} and {\acastle}, we consider two other baseline models. The first one optimizes the Eq.~\ref{eq:gfsl_meta_obj} directly but without the neural dictionary, which relies on both the (fixed) linear classifier $\mathbf{\Theta}_{\calS}$ and the few-shot prototypes to make a GFSL prediction (we denote it as ``\textsc{Castle}$^-$''); the second one is DFSL~\cite{Gidaris2018Dynamic}, which requires a two-stage training of the GFSL model. It trains a many-shot classifier with cosine similarity in the first stage. Then it freezes the backbone model as a feature extractor and optimizes a similar form of Eq.~\ref{eq:gfsl_meta_obj} via composing new few-shot classifiers as the convex combination of those many-shot classifiers. It can be viewed as a degenerated neural dictionary, where DFSL sets a size-$|\calS|$ ``shared'' bases $\mathcal{B}_{\mathbf{share}}$ as the many-shot classifier $\mathbf{\Theta}_{\calS}$. We observe that DFSL is unstable to perform end-to-end learning. It is potentially because the few-shot classifier composition uses many-shot classifiers as bases, but those bases are optimized to both be good bases and good classifiers, which can likely to be conflicting to some degree. It is also worth noting that all the baselines except {\acastle} only modify the few-shot classifiers, and it is impossible for them to perform backward knowledge transfer.

\subsection{GFSL over Heterogeneous Dataset}
The Heterogeneous dataset has 100 \textsc{seen} classes in the meta-training set, 20 per domain. We consider the case where during the inference, all of the tail classes come from one particular domain. For example, the tail classes are different kinds of birds, and we need to do a joint classification over all \textsc{seen} classes from the heterogeneous domains and the newly coming tail classes with limited instances. To mimic the inference scenario, we sample ``fake'' few-shot tasks with classes from one of the five domains randomly and contrasting the discerning ability from the sampled classes w.r.t. the remaining \textsc{seen} classes as in Eq.~\ref{eq:gfsl_meta_obj}. 

Note that we train DFSL strictly follows the strategy in~\cite{Gidaris2018Dynamic}, and train other GFSL models with the pre-trained embedding and the multi-classifier techniques to improve the training efficiency. 
Following~\cite{Xian2017Zero,Akata2018Generalized,Gidaris2018Dynamic}, we compute the 1-Shot 5-Way GFSL classification mean accuracy and harmonic mean accuracy over 10,000 sampled tasks, whose results are recorded in Table~\ref{tab:hetero_gfsl}.  $\calS \rightarrow \calS\cup\calU$ and  $\calU \rightarrow \calS\cup\calU$ denote the average accuracy for the joint prediction of \textsc{seen} and \textsc{unseen} instances respectively.

From the results in Table~\ref{tab:hetero_gfsl}, DFSL could not work well due to its fixed embedding and restricted bases. 
\textsc{Castle}$^-$ is able to balance the training accuracy of both \textsc{seen} and \textsc{unseen} classes benefited from the pre-train strategy and the unified learning objective, which achieves the highest joint classification performance over \textsc{unseen} classes. 
The discriminative ability is further improved with the help of the neural dictionary. {\castle} performs better than its degenerated version, which verifies the effectiveness of the learned neural bases.
The neural dictionary encodes the common characteristic among all classes for the GFSL classification, so that {\castle} gets better mean accuracy and harmonic mean accuracy than \textsc{Castle}$^-$. Since {\acastle} is able to adapt both many-shot and few-shot classifiers conditioned on the context of the tail instances, it obtains the best GFSL performance in this case. It is notable that {\acastle} gets much higher joint classification accuracy for \textsc{seen} classes than other methods, which validates its ability to adapt the many-shot classifier over the \textsc{seen} classes based on the context of tail classes.

\subsection{GFSL over Office-Home Dataset}
We also investigate the similar multi-domain GFSL classification task over the Office-Home dataset. However, in this case, a single class could belong to all three domains. We consider the scenario to classify classes in a single domain and the domain of the classes should be inferred from the limited tail instances. In other words, we train a GFSL model over 25 classes, and each class has 3 sets of instances corresponding to the three domains. In meta-training, a 25-way \textsc{seen} class classifier is constructed. During the inference, the model is provided by another 5-way 1-shot set of \textsc{unseen} class instances from one single domain. The model is required to output a joint classifier for test instances from the whole 30 classes whose domains are the same as the one in the \textsc{unseen} class set.

Towards such a multi-domain GFSL task, we train a GFSL model by keeping the instances in both the few-shot fake tail task and corresponding test set from the same domain. We use the same set of comparison methods and evaluation protocols with the previous subsection. The mean accuracy, harmonic mean accuracy, and the specific accuracy for \textsc{seen} and \textsc{unseen} classes are shown in Table~\ref{tab:office_gfsl}.

Due to the ambiguity of domains for each class, the GFSL classification over Office-Home gives rise to a more difficult problem, while the results in Table~\ref{tab:office_gfsl} reveal a similar trend with those in Table~\ref{tab:hetero_gfsl}.
Since for Office-Home a single GFSL model needs to make the joint prediction over classes from multiple domains conditioned on different configurations of the tail few-shot tasks, the stationary \textsc{seen} class classifiers are not suitable for the classification over different domains. In this case, {\acastle} still achieves the best performance over different GFSL criteria, and gets larger superiority margins with the comparison methods.

\section{Experiments on GFSL}
\label{sec:experiment}

In this section, we design experiments on benchmark datasets to validate the effectiveness of the {\castle} and {\acastle} in GFSL (\cf \S~\ref{sec:exp_gfsl}). 
After a comprehensive comparison with competitive methods using various protocols, we analyze different aspects of GFSL approaches, and we observe the post calibration makes the FSL methods strong GFSL baselines. We verify that {\castle}/{\acastle} learn a better calibration between \textsc{seen} and \textsc{unseen} classifiers, and the neural dictionary makes {\castle}/{\acastle} persist its high discerning ability with incremental tail few-shot instances. 
Finally, we show that {\castle}/{\acastle} also benefit standard FSL performances (\cf \S~\ref{sec:exp_fsl}).

\subsection{Datasets}
Two benchmark datasets are used in our experiments. The \textbf{\textit{Mini}ImageNet} dataset~\citep{VinyalsBLKW16Matching} is a subset of the ILSVRC-12 dataset~\citep{RussakovskyDSKS15ImageNet}. There are totally 100 classes and 600 examples in each class. For evaluation, we follow the split of~\citep{Sachin2017} and use 64 of 100 classes for meta-training, 16 for validation, and 20 for meta-test (model evaluation). In other words, a model is trained on few-shot tasks sampled from the 64 \textsc{seen} classes set during meta-training, and the best model is selected based on the few-shot classification performance over the 16 class set. The final model is evaluated based on few-shot tasks sampled from the 20 \textsc{unseen} classes.

The \textbf{\textit{Tiered}ImageNet}~\citep{Ren2018Meta} is a more complicated version compared with the {\it Mini}ImageNet. It contains 34 super-categories in total, with 20 for meta-training, 6 for validation (meta-val), and 8 for model testing (meta-test). Each of the super-category has 10 to 30 classes. In detail, there are 351, 97, and 160 classes for meta-training, meta-validation, and meta-test, respectively. The divergence of the super-concept leads to a more difficult few-shot classification problem. 

Since both datasets are constructed by images from ILSVRC-12, we augment the {\em meta-training} set of each dataset by sampling non-overlapping images from the corresponding classes in ILSVRC-12. The auxiliary meta-train set is used to measure the generalized few-shot learning classification performance on the \textsc{seen} class set. For example, for each of the 64 \textsc{seen} classes in the {\it Mini}ImageNet, we collect 200 more non-overlapping images per class from ILSVRC-12 as the test set for many-shot classification. The illustration of the dataset split is shown in Figure~\ref{fig:data_split}.

\subsection{Baselines and Prior Methods} 
We explore several (strong) choices in deriving classifiers for the \textsc{seen} and \textsc{unseen} classes, including Multiclass Classifier (MC) + $k$NN, which contains a $|\calS|$-way classifier trained on the \textsc{seen} classes in a supervised learning manner as standard many-shot classification, and its embedding with the nearest neighbor classifier is used for GFSL inference; ProtoNet + ProtoNet, where the embeddings trained by Prototypical Network~\citep{SnellSZ17Prototypical} is used, and 100 training instances are sampled from each \textsc{seen} category to act as the \textsc{seen} class prototypes; MC + ProtoNet, where we combine the learning objective of the previous two baselines to jointly learn the MC classifier and feature embedding. Details of the methods are in Appendix~\ref{sec:appendixa3}.

Besides, we also compare our approach with the L2ML~\cite{WangRH17Learning}, Dynamic Few-Shot Learning without forgetting~(\textbf{DFSL})~\citep{Gidaris2018Dynamic}, and the newly proposed Incremental few-shot learning~(\textbf{IFSL})~\citep{Ren2019Incremental}. For {\castle}, we use the  many-shot classifiers $\{\mathbf{\Theta}_\calS\}$ for the \textsc{seen} classes and the synthesized classifiers for the \textsc{unseen} classes to classify an instance into all classes, and then select the prediction with the highest confidence score. For {\acastle}, we adapt the head classifiers to $\{\hat{\mathbf{\Theta}}_\calS\}$ with the help of the tail classes.

\begin{table*}[t]
	\centering
	\small
	\tabcolsep 8pt
	\caption{
		Generalized Few-shot classification performance (mean accuracy, $\Delta$-value, and harmonic mean accuracy) on {\it Mini}ImageNet when there are {\bf 64 Head and 5 Tail categories}. 
	}
	{ 
		\small
		\begin{tabular}{@{\;}l@{\;}cccc|c@{\;}c}
			\addlinespace
			\toprule
			{Setups} & \multicolumn{ 2}{c}{1-Shot} & \multicolumn{ 2}{c|}{5-Shot} & 1-Shot & 5-Shot \\
			{Perf. Measures} & {\it Mean Acc.}$\;\mathbf{\Theta}parrow$   & $\Delta\;\downarrow$ & {\it Mean  Acc.}$\;\mathbf{\Theta}parrow$   & $\Delta\;\downarrow$ & \multicolumn{ 2}{c}{\it Harmonic Mean Acc.}$\;\mathbf{\Theta}parrow$ \\
			\midrule
			IFSL~\citep{Ren2019Incremental}  & 54.95{\tiny $\pm$0.30} & 11.84 & 63.04{\tiny $\pm$0.30} & 10.66 &  -   & - \\
			L2ML~\cite{WangRH17Learning} & 46.25{\tiny $\pm$0.04} & 27.49 & 45.81{\tiny $\pm$0.03} & 35.53 & 2.98{\tiny $\pm$0.06} & 1.12{\tiny $\pm$0.04} \\
			DFSL~\cite{Gidaris2018Dynamic} & 63.36{\tiny $\pm$0.11} & 13.71 & 72.58{\tiny $\pm$0.09} & 13.33 & 62.08{\tiny $\pm$0.13} & 71.26{\tiny $\pm$0.09} \\
			MC + $k$NN & 46.17{\tiny $\pm$0.03} & 29.70 & 46.18{\tiny $\pm$0.03} & 40.21 & 0.00{\tiny $\pm$0.00} & 0.00{\tiny $\pm$0.00} \\		
			MC + ProtoNet & 45.31{\tiny $\pm$0.03} & 29.71 & 45.85{\tiny $\pm$0.03} & 39.82 & 0.00{\tiny $\pm$0.00} & 0.00{\tiny $\pm$0.00} \\
			ProtoNet + ProtoNet & 50.49{\tiny $\pm$0.08} & 25.64 & 71.75{\tiny $\pm$0.08} & 13.65 & 19.26{\tiny $\pm$0.18} & 67.73{\tiny $\pm$0.12} \\
			\midrule
			\textbf{Ours:} {\castle} & 67.13{\tiny $\pm$0.11} & 10.09 & 76.78{\tiny $\pm$0.09} &  9.88 & 66.22{\tiny $\pm$0.15} &  76.32{\tiny $\pm$0.09} \\
			\textbf{Ours:} {\acastle} & \bf 68.70{\tiny $\pm$0.11} & \bf 9.98 & \bf 78.63{\tiny $\pm$0.09} & \bf 8.08 & \bf 66.24{\tiny $\pm$0.15} & \bf 78.33{\tiny $\pm$0.09} \\
			\bottomrule
	\end{tabular}}\label{tab:gfsl_acc}
\end{table*}

\subsection{Main Results}
\label{sec:exp_gfsl}

We first evaluate all GFSL methods on \textit{Mini}ImageNet with the criteria in~\citep{Gidaris2018Dynamic,Ren2019Incremental}, the mean accuracy over all classes (the higher the better) and the $\Delta$-value (the lower the better). An effective GFSL approach not only makes predictions well on the joint label space (with high accuracy) but also keeps its classification ability when changing from many-shot/few-shot to the generalized few-shot case (low $\Delta$-value).

The main results are shown in Table~\ref{tab:gfsl_acc}. 
We found that {\acastle} outperforms all the existing methods as well as our proposed baseline systems in terms of the mean accuracy.
Meanwhile, when looked at the $\Delta$-value, and {\castle} variants are the least affected between predicting for $\textsc{seen}$/$\textsc{usseen}$ classes separately and predicting over all classes jointly. 

However, we find that either mean accuracy or $\Delta$-value is not informative enough to tell about a GFSL algorithm's performance. For example, a baseline system, \ie., ProtoNet + ProtoNet performs better than IFSL in terms of 5-shot mean accuracy but not $\Delta$-value. This is consistent with the observation in~\cite{Ren2019Incremental} that the $\Delta$-value should be considered together with the mean accuracy.
\textit{In this case, how shall we rank these two systems?} To answer this question, we propose to use another evaluation measure, the harmonic mean of the mean accuracy for each \textsc{seen} and \textsc{unseen} category~\cite{Xian2017Zero,Akata2018Generalized}, when they are classified jointly. 


\mypara{Harmonic mean accuracy measures GFSL performance better.} 
Since the number of \textsc{seen} and \textsc{unseen} classes are most likely to be not equal, \eg., 64 vs. 5 in our cases, directly computing the mean accuracy over all classes is almost always biased. 
For example, a many-shot classifier that only classifies samples into \textsc{seen} classes can receive a good performance than one that recognizes both \textsc{seen} and \textsc{unseen}.
Therefore, we argue that \emph{harmonic mean} over the mean accuracy can better assess a classifier's performance, as now the performances are negatively affected when a classifier ignores classes (\eg., MC classifier get $0\%$ harmonic mean). 
Specifically, we compute the top-1 accuracy for instances from \textsc{seen} and \textsc{unseen} classes, and take their harmonic mean as the performance measure. 
The results are included in the right side of the Table~\ref{tab:gfsl_acc}. 

We find the harmonic mean accuracy takes a holistic consideration of the ``absolute'' joint classification performance and the ``relative'' performance drop when classifying towards the joint set. For example, the many-shot baseline MC+$k$NN with good mean accuracy and high $\Delta$-value has extremely low performance as it tends to ignore \textsc{unseen} categories. Meanwhile, {\castle} and {\acastle} remain the best when ranked by the harmonic mean accuracy against others. 

\begin{table*}[t]
	\centering
	\small
	\tabcolsep 2pt
	\caption{
		Generalized Few-shot classification accuracies on {\it Mini}ImageNet with \textbf{64 head categories and 20 tail categories}.
	}
	\resizebox{\linewidth}{!}
	{ \small
		\begin{tabular}{@{\;}l@{\;}cc@{\quad}cc@{\quad}cc@{\quad}cc@{\;}}
			\addlinespace
			\toprule
			Classification on  & \multicolumn{2}{c}{20 \textsc{unseen} Categories} & \multicolumn{6}{c}{64 \textsc{seen} + 20 \textsc{unseen} Categories } \\ 
			Perf. Measures  & \multicolumn{2}{c}{$\calU\rightarrow\calU$} & \multicolumn{2}{c}{$\calS\rightarrow\calS\cup\calU$} & \multicolumn{2}{c}{$\calU\rightarrow\calS\cup\calU$} & \multicolumn{2}{c}{\it HM Acc.} \\ \cmidrule(lr){2-3}\cmidrule(lr){4-9}
			Setups  & {1-Shot} & {5-Shot} & {1-Shot} & {5-Shot} & {1-Shot} & {5-Shot} & {1-Shot} & {5-Shot} \\
			\midrule
			L2ML~\cite{WangRH17Learning} & 27.79{\tiny $\pm$ 0.07} & 43.42{\tiny $\pm$ 0.06} &  90.99{\tiny $\pm$ 0.03} & 90.99{\tiny $\pm$ 0.03} & 0.64{\tiny $\pm$ 0.00} & 1.21{\tiny $\pm$ 0.01} & 1.27{\tiny $\pm$ 0.09} & 2.38{\tiny $\pm$ 0.02}  \\
			DFSL~\cite{Gidaris2018Dynamic} & 33.02{\tiny $\pm$ 0.08} & 50.96{\tiny $\pm$ 0.07} & 61.68{\tiny $\pm$ 0.06} & 66.06{\tiny $\pm$ 0.05} & \bf 31.13{\tiny $\pm$ 0.07} & \bf 47.16{\tiny $\pm$ 0.06} & 41.21{\tiny $\pm$ 0.07} & 54.95{\tiny $\pm$ 0.05}  \\
			MC + $k$NN & 31.58{\tiny $\pm$ 0.08} & 56.08{\tiny $\pm$ 0.06} & \bf 92.35{\tiny $\pm$ 0.03}  & 92.38{\tiny $\pm$ 0.03} & 0.00{\tiny $\pm$ 0.00} & 0.00{\tiny $\pm$ 0.00} & 0.00{\tiny $\pm$ 0.00} & 0.00{\tiny $\pm$ 0.00} \\
			MC + ProtoNet & 31.82{\tiny $\pm$ 0.06} & 56.16{\tiny $\pm$ 0.06} & 91.39{\tiny $\pm$ 0.03}  & \bf  92.99{\tiny $\pm$ 0.03} & 0.00{\tiny $\pm$ 0.00} & 0.00{\tiny $\pm$ 0.00} & 0.00{\tiny $\pm$ 0.00} & 0.00{\tiny $\pm$ 0.00}  \\
			ProtoNet + ProtoNet & 32.90{\tiny $\pm$ 0.08} & 55.69{\tiny $\pm$ 0.06} & 89.15{\tiny $\pm$ 0.04}  & 85.17{\tiny $\pm$ 0.04} & 9.89{\tiny $\pm$ 0.05} & 41.17{\tiny $\pm$ 0.06} & 17.71{\tiny $\pm$ 0.08} & 55.51{\tiny $\pm$ 0.06}\\
			\midrule
			\textbf{Ours:} {\castle} & {35.69{\tiny $\pm$ 0.08}} & {56.97{\tiny $\pm$ 0.06}} & 80.32{\tiny $\pm$ 0.06} & 80.43{\tiny $\pm$ 0.06}  & 29.42{\tiny $\pm$ 0.08} & 42.55{\tiny $\pm$ 0.05} & {43.06{\tiny $\pm$ 0.07}} & {55.65{\tiny $\pm$ 0.07}} \\
			\textbf{Ours:} {\acastle} & \textbf{36.38{\tiny $\pm$ 0.08}} & \textbf{57.29{\tiny $\pm$ 0.06}} & 81.36{\tiny $\pm$ 0.05}  & 87.40{\tiny $\pm$ 0.04} & 29.95{\tiny $\pm$ 0.08} & 41.64{\tiny $\pm$ 0.06} & \textbf{43.63{\tiny $\pm$ 0.08}} & \textbf{56.33{\tiny $\pm$ 0.06}} \\
			\bottomrule
	\end{tabular}}
	\label{tab:gfsl_miniimagenet}
\end{table*}

\begin{table*}[t]
	\small
	\centering
	\tabcolsep 2pt
	\caption{Generalized Few-shot classification accuracy on {\it Tiered}ImageNet with \textbf{351 head categories and 160 tail categories}.
	}
	\resizebox{\linewidth}{!}
	{\small
		\begin{tabular}{@{\;}l@{\;}cc@{\quad}cc@{\quad}cc@{\quad}cc@{\;}}
			\addlinespace
			\toprule
			Classification on  & \multicolumn{2}{c}{160 \textsc{unseen} Categories} & \multicolumn{6}{c}{351 \textsc{seen} + 160 \textsc{unseen} Categories } \\
			Perf. Measures  & \multicolumn{2}{c}{$\calU\rightarrow\calU$} & \multicolumn{2}{c}{$\calS\rightarrow\calS\cup\calU$} & \multicolumn{2}{c}{$\calU\rightarrow\calS\cup\calU$} & \multicolumn{2}{c}{\it HM Acc.} \\ \cmidrule(lr){2-3}\cmidrule(lr){4-9}
			Setups  & {1-Shot} & {5-Shot} & {1-Shot} & {5-Shot} & {1-Shot} & {5-Shot} & {1-Shot} & {5-Shot} \\
			\midrule
			DFSL~\cite{Gidaris2018Dynamic} & 15.79{\tiny $\pm$ 0.02} & 30.69{\tiny $\pm$ 0.02} & 11.29{\tiny $\pm$ 0.05} & 14.95{\tiny $\pm$ 0.06} & 14.24{\tiny $\pm$ 0.06} & 27.22{\tiny $\pm$ 0.07} & 12.60{\tiny $\pm$ 0.11} & 19.29{\tiny $\pm$ 0.05}  \\
			MC + $k$NN  & 14.12{\tiny $\pm$ 0.02} & 30.02{\tiny $\pm$ 0.02} & 68.32{\tiny $\pm$ 0.02} & \bf 68.33{\tiny $\pm$ 0.02} & 0.00{\tiny $\pm$ 0.00}  & 0.00{\tiny $\pm$ 0.00} & 0.01{\tiny $\pm$ 0.00} & 0.01{\tiny $\pm$ 0.00} \\
			MC + ProtoNet  & 14.13{\tiny $\pm$ 0.02} & 30.05{\tiny $\pm$ 0.02} & \bf 68.34{\tiny $\pm$ 0.02} & \bf 68.33{\tiny $\pm$ 0.02} & 0.00{\tiny $\pm$ 0.00} & 0.00{\tiny $\pm$ 0.00} & 0.00{\tiny $\pm$ 0.00} & 0.00{\tiny $\pm$ 0.00}  \\
			ProtoNet + ProtoNet    & 14.52{\tiny $\pm$ 0.02} & 29.38{\tiny $\pm$ 0.02} & 62.37{\tiny $\pm$ 0.02} & 61.15{\tiny $\pm$ 0.02} & 4.83{\tiny $\pm$ 0.03} & 22.69{\tiny $\pm$ 0.02} & 8.97{\tiny $\pm$ 0.02} & 33.09{\tiny $\pm$ 0.02}\\
			\midrule
			\textbf{Ours:} {\castle} &  15.97{\tiny $\pm$ 0.02} & 30.44{\tiny $\pm$ 0.02} & 26.94{\tiny $\pm$ 0.08} & 34.98{\tiny $\pm$ 0.02} & \bf 16.17{\tiny $\pm$ 0.06} & 31.61{\tiny $\pm$ 0.05} & 20.20{\tiny $\pm$ 0.05} & 33.20{\tiny $\pm$ 0.02} \\
			\textbf{Ours:} {\acastle} & \bf 16.36{\tiny $\pm$ 0.02} & \bf 30.75{\tiny $\pm$ 0.02} & 27.01{\tiny $\pm$ 0.08} & 35.41{\tiny $\pm$ 0.08} & \bf 16.17{\tiny $\pm$ 0.06} & \bf 31.86{\tiny $\pm$ 0.05} & \bf 22.23{\tiny $\pm$ 0.05} & \bf 33.54{\tiny $\pm$ 0.02} \\
			\bottomrule
	\end{tabular}}
	\label{tab:gfsl_tieredimagenet}
\end{table*}

\mypara{Evaluate GFSL beyond 5 \textsc{unseen} categories.} Besides using harmonic mean accuracy, we argue that another important aspect in evaluating GFSL is to go beyond the 5 sampled \textsc{unseen} categories, as it is never the case in real-world. On the contrary, we care most about the GFSL with a large number of \textsc{unseen} classes, which also measure the ability of the model to extrapolate the number of novel classes in the \textsc{unseen} class few-shot task. To this end, we consider an extreme case -- evaluating GFSL with {\em all available} \textsc{seen} and \textsc{unseen} categories over both {\it Mini}ImageNet and {\it Tiered}ImageNet, and report their results in Table~\ref{tab:gfsl_miniimagenet} and Table~\ref{tab:gfsl_tieredimagenet}.

Together with the harmonic mean accuracy of \emph{all} categories, we also report the tail classification performance, which is a more challenging few-shot classification task (the standard FSL results could be found in \S~\ref{sec:exp_fsl}). In addition, the joint classification accuracy for \textsc{seen} classes instances ($\calS\rightarrow\calS\cup\calU$) and \textsc{unseen} classes instances ($\calU\rightarrow\calS\cup\calU$) are also listed.

The methods without a clear consideration of head-tail trade-off (\eg., ProtoNet+ProtoNet) fails to make a joint prediction over both \textsc{seen} and \textsc{unseen} classes.
We observe that {\castle} and {\acastle} outperform all approaches in the \textsc{unseen} and more importantly, the \textsc{all} categories section, across two datasets. 

\begin{figure*}[tbp]
	\centering
	\tabcolsep 3pt
	\small
	\begin{tabular}{ccc}
		\begin{minipage}[h]{0.47\linewidth}
			\includegraphics[width=1.0\textwidth]{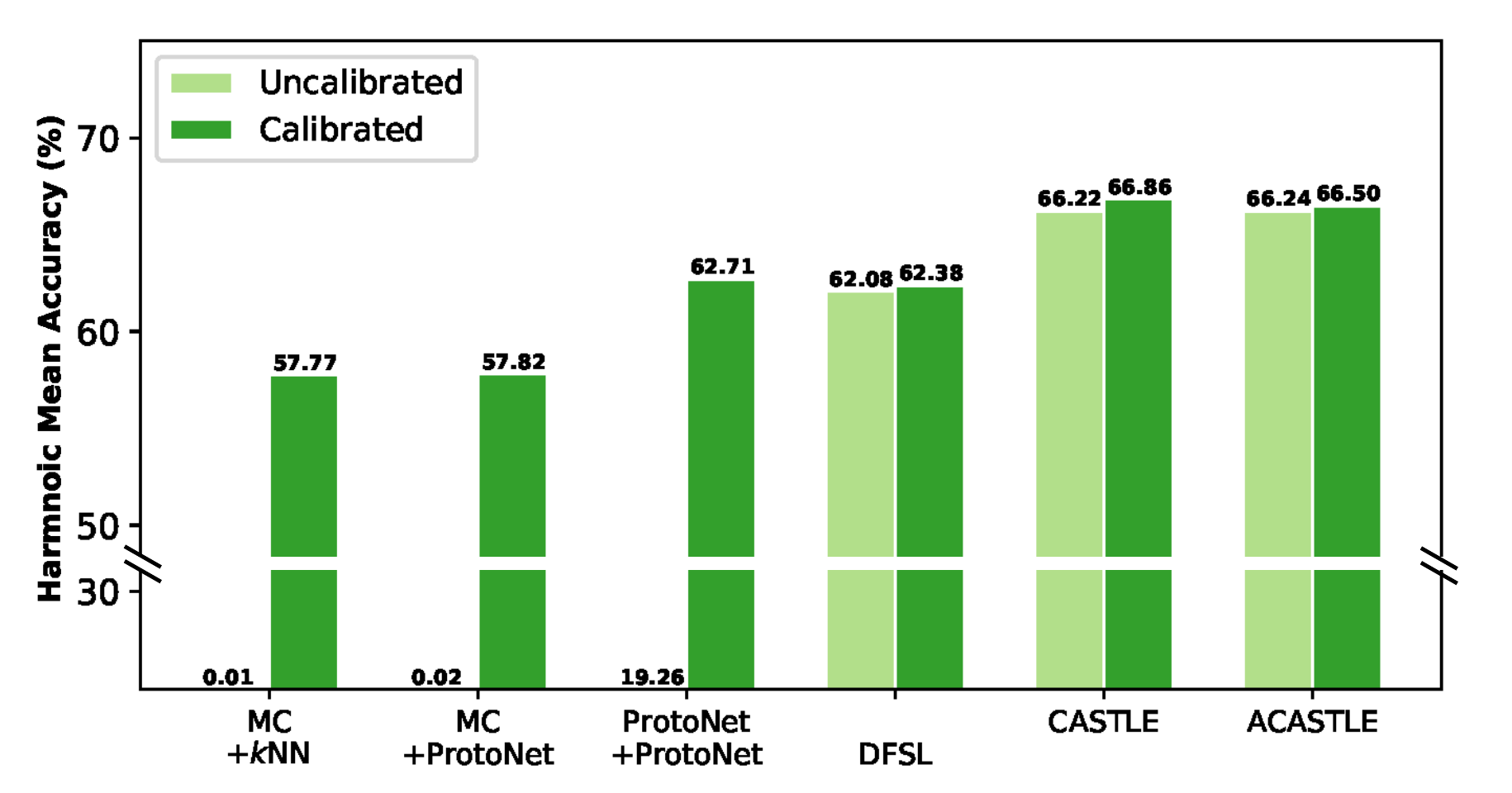}
			\caption{Calibration's effect to the 1-shot harmonic mean accuracy on {\it Mini}ImageNet. Baseline models improve a lot with the help of the calibration factor.}
			\label{fig:calibration}
		\end{minipage}
		&
		\begin{minipage}[h]{0.47\linewidth}
			\includegraphics[width=1.0\textwidth]{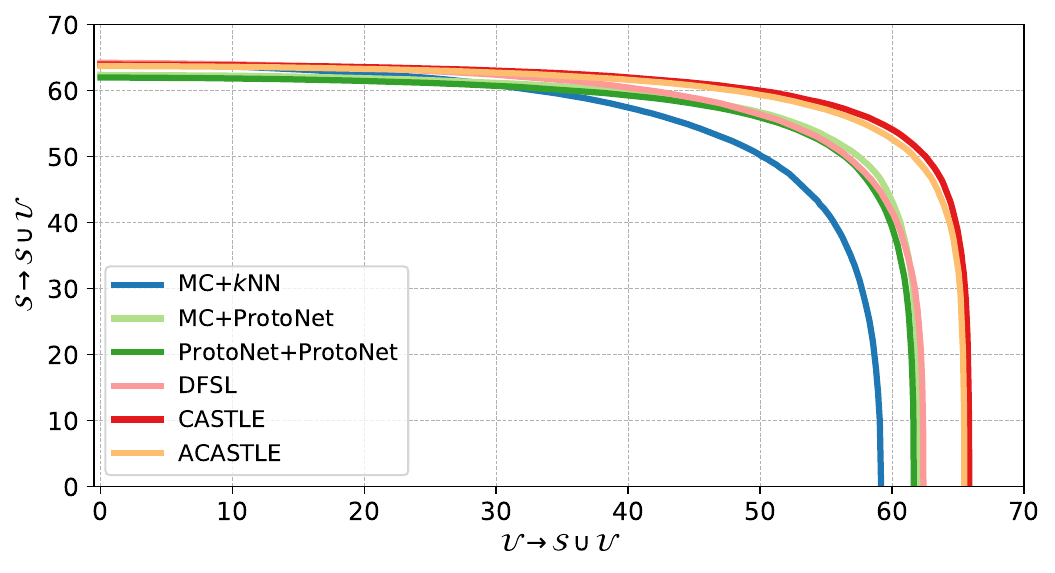}
			\caption{The 1-shot AUSUC performance with two configurations of \textsc{unseen} classes on {\it Mini}ImageNet. The larger the area under the curve, the better the GFSL ability.}
			\label{fig:AUSUC_1_shot}
		\end{minipage}
	\end{tabular}
\end{figure*}

\mypara{Confidence calibration matters in GFSL.} In generalized zero-shot learning, Chao~\etal~\cite{Chao2016Generalized} have identified a significant prediction bias between classification confidence of \textsc{seen} and \textsc{unseen} classifiers. We find a similar phenomena in GFSL. For instance, the few-shot learning \textit{ProtoNet + ProtoNet} baseline becomes too confident to predict on \textsc{seen} categories than \textsc{unseen} categories (The scale of confidence is on average 2.1 times higher).
To address this issue, we compute a calibration factor based on the meta-validation set of \textsc{unseen} categories, such that the prediction logits are calibrated by subtracting this factor out from the confidence of \textsc{seen} categories' predictions.
With 5 \textsc{unseen} classes from {\it Mini}ImageNet, the GFSL results of all comparison methods before and after calibration is shown in Figure~\ref{fig:calibration}. 
We observe consistent and obvious improvements over the harmonic mean accuracy for all methods. For example, although the FSL approach ProtoNet neglects the classification performance over \textsc{seen} categories outside the sampled task during meta-learning, it gets even better harmonic mean accuracy compared with the GFSL method DFSL (62.70\% vs. 62.38\%) with such post-calibration, which {\em becomes a very strong GFSL baseline}.
Note that {\castle} and {\acastle} are the least affected with the selected calibration factor. 
This suggests that {\castle} variants, learned with the unified GFSL objective, have well-calibrated classification confidence and do not require additional data and extra learning phases to search this calibration factor.

\begin{figure*}[tbp]
	\centering
	\small
	\begin{tabular}{cc}
		\begin{minipage}[h]{0.47\linewidth}
			\tabcolsep 2pt
			\includegraphics[width=1.0\textwidth]{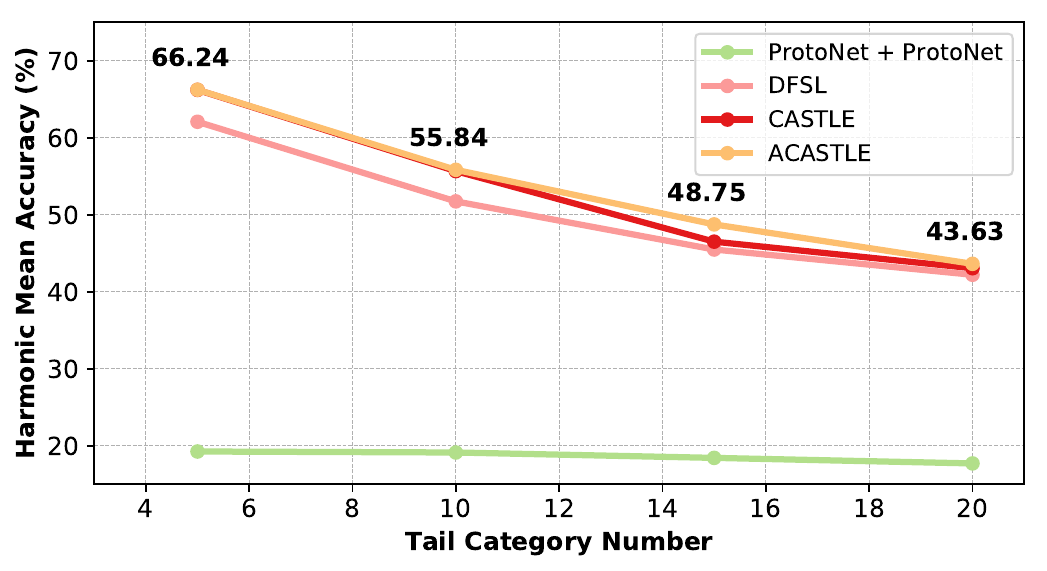}
			\caption{Results of 1-shot GFSL harmonic mean accuracy with incremental number of \textsc{unseen}  classes on {\it Mini}ImageNet. Note MC+$k$NN and MC+ProtoNet bias towards \textsc{seen} classes and get nearly zero harmonic mean accuracy.}
			\label{fig:incremental_1shot}
		\end{minipage}
		&
		\begin{minipage}[h]{0.47\linewidth}
			\includegraphics[width=1.0\textwidth]{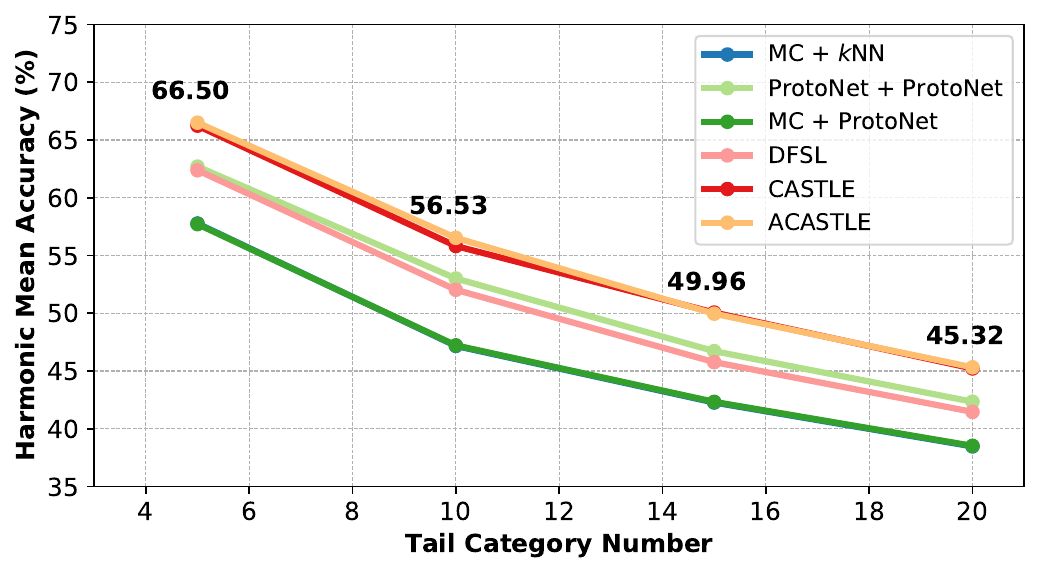}
			\caption{{\em Post-calibrated} results of 1-shot GFSL harmonic mean accuracy with incremental number of \textsc{unseen} classes on {\it Mini}ImageNet. All methods select the their best calibration factors from the meta-val data split.}
			\label{fig:incremental_1shot_cal}
		\end{minipage}
	\end{tabular}
\end{figure*}

Moreover, we use area under \textsc{seen}-\textsc{unseen} curve (AUSUC) as a measure of different GFSL algorithms~\cite{Chao2016Generalized}. Here, AUSUC is a performance measure that takes the effects of the calibration factor out. 
To do so, we enumerate through a large range of calibration factors and subtract it from the confidence score of \textsc{seen} classifiers. 
Through this process, the joint prediction performances over \textsc{seen} and \textsc{unseen} categories, denoted as $\calS\rightarrow \calS\cup\calU$ and $\calU\rightarrow \calS\cup\calU$, shall vary as the calibration factor changes. 
For instance, when the calibration factor is infinitely large, we measure a classifier that only predicts \textsc{unseen} categories. We denote this as the \textsc{seen}-\textsc{unseen} curve. The 1-shot GFSL results with 5 \textsc{unseen} classes from {\it Mini}ImageNet is shown in Figure~\ref{fig:AUSUC_1_shot}. As a result, we observe that {\acastle} and {\castle} archive the largest area under the curve, which indicates that {\castle} variants are in general a better algorithm over others among different calibration factors.

\mypara{Robust evaluation of GFSL.} Other than the harmonic mean accuracy of all \textsc{seen} and \textsc{unseen} categories shown in Table~\ref{tab:gfsl_miniimagenet} and Table~\ref{tab:gfsl_tieredimagenet}, we study the dynamic of how harmonic mean accuracy changes with an incremental number of \textsc{unseen} tail concepts. 
In other words, we show the GFSL performances \wrt. different numbers  of tail concepts. We use this as a {\em robust evaluation} of each system's GFSL capability. 
In addition to the test instances from the head 64 classes in {\it Mini}ImageNet, 5 to 20 novel classes are included to compose the generalized few-shot tasks. Concretely, only one instance per novel class is used to construct the tail classifier, combined with which the model is asked to do a {\em joint} classification of both \textsc{seen} and \textsc{unseen} classes.  Figure~\ref{fig:incremental_1shot} records the change of generalized few-shot learning performance (harmonic mean) when more \textsc{unseen} classes emerge. We omit the results of MC+$k$NN and MC+ProtoNet since they bias towards \textsc{seen} classes and get nearly zero harmonic mean accuracy in all cases. We observe that \acastle consistently outperforms all baseline approaches in each evaluation setup, with a clear margin. We also compute the harmonic mean after selecting the best calibration factor from the meta-val set (\cf Figure~\ref{fig:incremental_1shot_cal}). It is obvious that almost all baseline models achieve improvements and the phenomenon is consistent with Figure~\ref{fig:calibration}. The GFSL results of \acastle and \castle are almost not influenced after using the post-calibration technique. \acastle still persists its superiority in this case.


\subsection{Standard Few-Shot Learning}
\label{sec:exp_fsl}
\begin{table*}[tbp]
    \centering
    \begin{tabular}{cc}
        \begin{minipage}[h]{0.47\linewidth}
        	\tabcolsep 8pt
        	\scriptsize
            \caption{Few-shot classification accuracy on {\it Mini}ImageNet with different types of backbones. Our methods are evaluated with 10,000 few-shot tasks.}
            \resizebox{\linewidth}{!}{ 
	            \begin{tabular}{@{\;}l@{\;}c@{\;}c@{\;}c@{\;}}
	            \addlinespace
	            \toprule
	            Setups & Backbone & {\bf 1-Shot 5-Way} & {\bf 5-Shot 5-Way} \\
	            \midrule
	            IFSL~\citep{Ren2019Incremental} & ResNet-10 &  55.72 {\tiny $\pm$ 0.41} &  70.50 {\tiny $\pm$ 0.36} \\
	            DFSL~\cite{Gidaris2018Dynamic} & ResNet-10 &  56.20 {\tiny $\pm$ 0.86} &  73.00 {\tiny $\pm$ 0.64} \\
	            ProtoNet~\citep{SnellSZ17Prototypical} & ResNet-12 & 61.40 {\tiny $\pm$ 0.12} & 76.56 {\tiny $\pm$ 0.20} \\
	            TapNet~\citep{Yoon2019Tap} & ResNet-12 &  61.65 {\tiny $\pm$ 0.15} &  76.36 {\tiny $\pm$ 0.10} \\
	            MTL~\citep{Sun2019Meta} & ResNet-12 &  61.20 {\tiny $\pm$ 1.80} &  75.50 {\tiny $\pm$ 0.90} \\
	            MetaOptNet~\citep{Lee2019Meta} & ResNet-12 &  62.64 {\tiny $\pm$ 0.61} &  78.63 {\tiny $\pm$ 0.46} \\
	            \textsc{Feat}~\citep{YeHZS2018Learning} & ResNet-12 &  66.78 {\tiny $\pm$ 0.20} &  82.05 {\tiny $\pm$ 0.14} \\
	           	SimpleShot~\citep{Wang2019Simple} & ResNet-18 &  62.85 {\tiny $\pm$ 0.20} &  80.02 {\tiny $\pm$ 0.14} \\
	           	CTM~\citep{Li2019Finding} & ResNet-18 &  64.12 {\tiny $\pm$ 0.82} &  80.51 {\tiny $\pm$ 0.13} \\
	            LEO~\citep{Rusu2018Meta} & WRN & 61.76 {\tiny $\pm$ 0.08} & 77.59 {\tiny $\pm$ 0.12} \\
	            \midrule
	            \textbf{Ours:} {\castle} & ResNet-12 & 66.75 {\tiny $\pm$ 0.20} & {81.98} {\tiny $\pm$ 0.14} \\
	            \textbf{Ours:} {\acastle} & ResNet-12 & {\bf 66.83} {\tiny $\pm$ 0.20} & \bf {82.08} {\tiny $\pm$ 0.14} \\
	             \bottomrule
	            \end{tabular}        
    		}
            \label{tab:miniImageNet}
        \end{minipage}
        &
        \begin{minipage}[h]{0.47\linewidth}
        	\tabcolsep 8pt
        	\scriptsize
            \caption{Few-shot classification accuracy on {\it Tiered}ImageNet with different types of backbones. Our methods are evaluated with 10,000 few-shot tasks.}
            \resizebox{\linewidth}{!}{
	            \begin{tabular}{@{\;}l@{\;}c@{\;}c@{\;}c@{\;}}
	            \addlinespace
	            \toprule
	            Setups & Backbone & {\bf 1-Shot 5-Way} & {\bf 5-Shot 5-Way} \\
	            \midrule
	            ProtoNet~\citep{SnellSZ17Prototypical} & ConvNet & 53.31 {\tiny $\pm$ 0.89} & 72.69 {\tiny $\pm$ 0.74} \\
	            IFSL~\citep{Ren2019Incremental} & ResNet-18 &  51.12 {\tiny $\pm$ 0.45} &  66.40 {\tiny $\pm$ 0.36} \\
	            DFSL~\cite{Gidaris2018Dynamic} & ResNet-18 &  50.90 {\tiny $\pm$ 0.46} &  66.69 {\tiny $\pm$ 0.36} \\
	            TapNet~\citep{Yoon2019Tap} & ResNet-12 &  63.08 {\tiny $\pm$ 0.15} &  80.26 {\tiny $\pm$ 0.12} \\
	            MTL~\citep{Sun2019Meta} & ResNet-12 &  65.60 {\tiny $\pm$ 1.80} &  78.60 {\tiny $\pm$ 0.90} \\
	            MetaOptNet~\citep{Lee2019Meta} & ResNet-12 &  65.99 {\tiny $\pm$ 0.72} &  81.56 {\tiny $\pm$ 0.63} \\
	            \textsc{Feat}~\citep{YeHZS2018Learning} & ResNet-12 &  70.80 {\tiny $\pm$ 0.23} &  84.79 {\tiny $\pm$ 0.16} \\
	            SimpleShot~\citep{Wang2019Simple} & ResNet-18 &  69.09 {\tiny $\pm$ 0.22} &  84.58 {\tiny $\pm$ 0.16} \\
	            CTM~\citep{Li2019Finding} & ResNet-18 &  68.41 {\tiny $\pm$ 0.39} &  84.28 {\tiny $\pm$ 1.73} \\
	           	LEO~\citep{Rusu2018Meta}  & WRN & 66.33 {\tiny $\pm$ 0.05} & 81.44 {\tiny $\pm$ 0.09} \\
	            \midrule
	            \textbf{Ours:} {\castle} & ResNet-12 & 71.14 {\tiny $\pm$ 0.02} & {84.34} {\tiny $\pm$ 0.16} \\
	            \textbf{Ours:} {\acastle} & ResNet-12 & {\bf 71.63} {\tiny $\pm$ 0.02} & \bf {85.28} {\tiny $\pm$ 0.15} \\
	             \bottomrule
	            \end{tabular}
	            \label{tab:tieredImageNet}
        	}
        \end{minipage}
    \end{tabular}
\end{table*}

Finally, we also evaluate our proposed approaches' performance on two standard few-shot learning benchmarks, \ie., {\it Mini}ImageNet and {\it Tiered}ImageNet dataset. In other words, we evaluate the classification performance of few-shot \textsc{unseen} class instances with our GFSL objective. We compare our approaches with the state-of-the-art methods in both 1-shot 5-way and 5-shot 5-way scenarios. We cite the results of the comparison methods from their published papers and remark the backbones used to train the FSL model by different methods. The mean accuracy and 95\% confidence interval are shown in the Table~\ref{tab:miniImageNet} and Table~\ref{tab:tieredImageNet}. 

It is notable that some comparison methods such as CTM~\cite{Li2019Finding} are evaluated over only 600 \textsc{unseen} class FSL tasks, while we test both {\castle} and {\acastle} over 10,000 tasks, leading to more stable results. {\castle} and {\acastle} achieve almost the best 1-shot and 5-shot classification results on both datasets.
The results support our hypothesis that jointly learning with many-shot classification forces few-shot classifiers to be discriminative.

\section{Related Work and Discussion}
\label{sec:related}
Building a high-quality classification system usually requires to have a large scale of annotated training set with many shots per category. Many large-scale datasets such as ImageNet have an ample number of instances for popular classes~\cite{RussakovskyDSKS15ImageNet,Krizhevsky2017ImageNet}. However, the data-scarce tail of the category distribution matters. For example, a visual search engine needs to deal with the rare object of interests (\eg., endangered species) or newly defined items (\eg., new smartphone models), which only possesses a few data instances.
Directly training a system over all classes is prone to over-fit and can be biased towards the data-rich categories~\cite{Cui2019Class,Cao2019Margin,Kang2020Decoupling,Ye2020Identifying,Zhou2020BBN}. 

Zero-shot learning (ZSL)~\cite{lampert2014attribute,akata2013label,Xian2017Zero,Changpinyo2018Classifier} is a popular idea for addressing learning without labeled data. It transfers the relationship between images and attributes learned from \textsc{seen} classes to \textsc{unseen} classes, using the semantic descriptions of objects as a bridge. For instance, popular methods~\cite{Changpinyo2018Classifier,changpinyo2017predicting} learn a mapping from the semantic descriptions (\eg, word embedings of category name) to its corresponding visual prototype. As a result, one can infer the visual prototype of unseen classes by looking at its corresponding semantic descriptions. 
Both ZSLs are limited to recognizing objects with well-defined semantic descriptions, which assumes that the visual appearance of novel categories is harder to obtain than knowledge about their attributes, whereas in the real-world we often get the appearance of objects before learning about their characteristics.

Few-shot learning (FSL) proposes a more realistic setup, where we have access to a limited number (\eg, one example) of visual exemplars from the tail classes~\cite{Fei-FeiFP06One,VinyalsBLKW16Matching} in the deployment of the visual system, and are required to recognize new instances of these tail categories. 
It places the challenge asking for a classification model to rapidly pick up the key characteristic of those few training examples, and use them to build effective classifiers. 

To deal with this challenges, FSL algorithms typically simulates the learning situation they encountering in the deployment time, by using the training data of \textsc{seen} classes. Specifically, works uses meta-learning algorithms~\cite{FinnAL17Model, SnellSZ17Prototypical} to extract the inductive bias from the \textsc{seen} classes, and transfer it to the learning process of \textsc{unseen} classes with few training data during the model deployment. For example, one line of works uses meta-learned discriminative feature embeddings~\cite{SnellSZ17Prototypical,VinyalsBLKW16Matching} together with the non-parametric nearest neighbor classifiers agnostic to its context, to recognize novel classes given a few exemplars. Another line of works chooses to learn the common optimization strategy~\cite{Sachin2017,Bertinetto2019Meta} across few-shot tasks. Such strategy adapts a pre-specified model initialization to the context of the specific classification task, using gradient descents over the few-shot \textsc{unseen} training data~\cite{FinnAL17Model,LiZCL17Meta,Nichol2018On,Lee2018Gradient,Antoniou2018How}. 

Besides FSL, generalized few-shot learning (GFSL) takes a further step towards real-world usage, where a model is required to master the recognition of not only \textsc{unseen} tail classes but also \textsc{seen} head classes. As a result, generalized few-shot learning exposes \emph{two additional challenges}. First, an algorithm needs to construct classifiers not only for the few-shot tail classes but also for the many-shot head classes. More importantly, the learning of two types of classifiers needs to be integrated together such that the predictions are compatible to each other, without sacrificing recalls on either head or tail classes. 

Some of the previous GFSL approaches~\citep{Hariharan2017Low,Wang2018Low,Gao2018Low} apply the exemplar-based classification paradigms on both \textsc{seen} and \textsc{unseen} categories to resolve the transductive learning problem, which requires recomputing the centroids for \textsc{seen} categories after model updates. 
Others~\citep{WangRH17Learning,Akata2018Generalized,Liu2019Large} usually ignore the explicit relationship between \textsc{seen} and \textsc{unseen} categories, and learn separate classifiers.
\citep{Ren2019Incremental, Gidaris2018Dynamic} propose to solve inductive GFSL via either composing \textsc{unseen} with \textsc{seen} classifiers or meta-leaning with recurrent back-propagation procedure. 
Gidaris~\etal~\citep{Gidaris2018Dynamic} is the most related work to {\castle} and {\acastle}, which composes the tail classifiers by a convex combination of the many-shot classifiers. 
{\castle} is different from Gidaris~\etal~\citep{Gidaris2018Dynamic} as it presents an \textit{end-to-end learnable framework} with improved training techniques, as well as it employs \textit{a shared neural dictionary} to compose few-shot classifiers.
Moreover, {\acastle} further relates the knowledge for both \textsc{seen} and \textsc{unseen} classes by constructing a neural dictionary with both shared (yet task-agnostic) and task-specific basis, which allows backward knowledge transfer to benefit \textsc{seen} classifiers with new knowledge of \textsc{unseen} classes. As we have demonstrated in \S~\ref{sec:exp_cross}, {\acastle} significantly improves \textsc{seen} classifiers when learning of \textsc{unseen} visual categories over heterogeneous visual domains.

\section{Conclusion}
A Generalized Few-Shot Learning (GFSL) model takes both the discriminative ability of many-shot and few-shot classifiers into account.  In this paper, we propose the ClAssifier SynThesis LEarning (\castle) and its adaptive variant (\acastle) to solve the challenging inductive modeling of \textsc{unseen} tail categories in conjunction with \text{seen} head ones. 
Our approach takes advantage of the neural dictionary to learn bases for composing many-shot and few-shot classifiers via a unified learning objective, which transfers the knowledge from \textsc{seen} to \textsc{unseen} classifiers better. 
Our experiments highlight \acastle especially fits the GFSL scenario with tasks from multiple domains. Both \castle and \acastle not only outperform existing methods in terms of various GFSL criteria but also improve the classifier's discernibility over standard FSL.
Future directions include improving the architecture of the neural dictionary and designing better fine-tuning strategies for GFSL.

\begin{acknowledgements}
Thanks to Fei Sha for valuable discussions. 
This research is partially supported by the NSFC (61773198, 61751306, 61632004), NSFC-NRF Joint Research Project under Grant 61861146001, NSF Awards IIS-1513966/ 1632803/1833137, CCF-1139148, DARPA Award\#: FA8750-18-2-0117,  DARPA-D3M - Award UCB-00009528, Google Research Awards, gifts from Facebook and Netflix, and ARO\# W911NF-12-1-0241 and W911NF-15-1-0484.
\end{acknowledgements}

\setcounter{subsection}{0}
\section*{Appendix A Details of GFSL Baselines}
\renewcommand{\thesection}{A}
\label{sec:appendixa3}
Here we describe some baseline approaches compared in the GFSL benchmarks in detail.

\mypara{(1) Multiclass Classifier (MC) + $k$NN.} A $|\calS|$-way classifier is trained on the \textsc{seen} classes in a supervised learning manner as standard many-shot classification~\citep{he2016deep}. During the inference, test examples of $\calS$ categories are evaluated based on the $|\calS|$-way classifiers and $|\calU|$ categories are evaluated using the support embeddings from $\calD_{\mathbf{train}}^{\;\calU}$ with a nearest neighbor classifier. To evaluate the generalized few-shot classification task, we take the union of multi-class classifiers' confidence and nearest neighbor confidence (the normalized negative distance values as in~\citep{SnellSZ17Prototypical}) as joint classification scores on $\calS \cup \calU$.

\mypara{(2) ProtoNet + ProtoNet.} We train a few-shot classifier (initialized by the MC classifier’s feature mapping) using the Prototypical Network~\citep{SnellSZ17Prototypical} (a.k.a. ProtoNet), pretending they were few-shot. When evaluated on the \textsc{seen} categories, we randomly sample 100 training instances per category to compute the class prototypes. The class prototypes of \textsc{unseen} classes are computed based on the sampled few-shot training set.
During the inference of \textit{generalized} few-shot learning, the confidence of a test instances is jointly determined by its (negative) distance to both \textsc{seen} and \textsc{unseen} class prototypes.

\mypara{(3) MC + ProtoNet.} We combine the learning objective of the previous two baselines ((1) and (2)) to jointly learn the MC classifier and feature embedding. Since there are two objectives for many-shot (cross-entropy loss on all \textsc{seen} classes) and few-shot (ProtoNet meta-learning objective) classification respectively, it trades off between  many-shot and few-shot learning. Therefore, this learned model can be used as multi-class linear classifiers on the head categories, and used as ProtoNet on the tail categories. During the inference, the model predicts instances from \textsc{seen} class $\calS$ with the MC classifier, while takes advantage of the few-shot prototypes to discern \textsc{unseen} class instances. To evaluate the generalized few-shot classification task, we take the union of multi-class classifiers' confidence and ProtoNet confidence as joint classification scores on $\calS \cup \calU$. 

\mypara{(4) L2ML.}
Wang \etal.~\cite{WangRH17Learning} propose learning to model the ``tail'' (L2ML) by connecting a few-shot classifier with the corresponding many-shot classifier. The method is designed to learn classifier dynamics from data-poor ``tail'' classes to the data-rich head classes.
Since L2ML is originally designed to learn with both \textsc{seen} and \textsc{unseen} classes in a transductive manner. In our experiment, we adaptive it to our setting. Therefore, we learn a classifier mapping based on the sampled few-shot tasks from \textsc{seen} class set $\calS$, which transforms a few-shot classifier in \textsc{unseen} class set $\calU$ inductively. Following~\citep{WangRH17Learning}, we first train a many-shot classifier $W$ upon the ResNet backbone on the \textsc{seen} class set $\calS$. We use the same residual architecture as in~\citep{WangRH17Learning} to implement the classifier mapping $f$, which transforms a few-shot classifier to a many-shot classifier. During the meta-learning stage, a $\calS$-way few-shot task is sampled in each mini-batch, which produces a $\calS$-way linear few-shot classifier $\hat{W}$ based on the fixed pre-trained embedding. The objective of L2ML not only regresses the mapped few-shot classifier $f(\hat{W})$ close to the many-shot one $W$ measured by square loss, but also minimizes the classification loss of $f(\hat{W})$ over a randomly sampled instances from $\calS$. 
Therefore, L2ML uses a pre-trained multi-class classifier $W$ for those head categories and used the predicted few-shot classifiers with $f$ for the tail categories.

\setcounter{subsection}{0}
\section*{Appendix B More Analysis on GFSL Benchmarks}
\renewcommand{\thesection}{B}
In this appendix, we do analyses to show the influence of training a GFSL model by reusing the many-shot classifier and study different implementation choices in the proposed methods. We mainly investigate and provide the results over {\castle} on {\it Mini}ImageNet. We observe the results on {\acastle} and other datasets reveal similar trends.

\begin{table}[t]
	\tabcolsep 5pt
	\centering
	\caption{The difference between training with a pre-trained backbone or from scratch with 1-Shot 5-Way Tasks on {\it Mini}ImageNet. ``MA'' and ``HM'' denote the {\it Mean Accuracy} and {\it Harmonic Mean Accuracy}, respectively. }
	\begin{tabular}{lcc}
		\addlinespace
		\toprule
		Perf. Measures  & FSL {\it MA}  & GFSL {\it HM} \\
		\midrule
		{\castle} w/ pre-train & 66.83 {\tiny $\pm$0.21} & 66.22 {\tiny $\pm$0.15} \\
		{\castle} w/o pre-train & 64.23 {\tiny $\pm$0.21} & 38.24 {\tiny $\pm$0.09}\\
		\bottomrule
	\end{tabular}
	\label{tab:pretrain}
\end{table}

\subsection{Reusing the many-shot classifier facilitates the calibration for GFSL}\label{sec:appendix-reuse}
We compare the strategy to train {\castle} from scratch and fine-tune based on the many-shot classifier. We show both the results of 1-Shot 5-Way few-shot classification performance and GFSL performance with 5 \textsc{unseen} tasks for {\castle} when trained from random or with provided initialization. From the results in Table~\ref{tab:pretrain}, we find training from scratch gets only a bit lower few-shot classification results with the fine-tune strategy, but much lower GFSL harmonic mean accuracy. Therefore, reusing the parameters in the many-shot classifier benefits the predictions on \textsc{seen} and \textsc{unseen} classes of a GFSL model. Therefore, we use the pre-trained embedding to initialize the backbone.

\begin{table}[t]
	\tabcolsep 5pt
	\centering
	\caption{Comparison between \castle variants and the incremental learning methods on {\it Mini}ImageNet. The harmonic mean accuracy in different evaluation scenarios are recorded.}
	\begin{tabular}{@{\;}l@{\;}c@{\;}c@{\;}c@{\;}c@{\;}}
		\addlinespace
		\toprule
		{Classification on} & \multicolumn{ 2}{c}{5-Way} & \multicolumn{ 2}{c}{20-Way} \\
		{Setups} & 1-Shot & 5-Shot & 1-Shot & 5-Shot\\
		\midrule
		LwF~\cite{Li2018Learning} & 60.18{\tiny $\pm$0.15} &  73.48{\tiny $\pm$0.09} & 28.70 {\tiny $\pm$0.06} & 39.88 {\tiny $\pm$0.06} \\
		iCARL\cite{Li2018Learning} & 61.14{\tiny $\pm$0.15} &  73.58{\tiny $\pm$0.09} & 31.60 {\tiny $\pm$0.06} & 46.55 {\tiny $\pm$0.06} \\
		{\castle} & 66.22{\tiny $\pm$0.15} &  76.32{\tiny $\pm$0.09} & {43.06{\tiny $\pm$ 0.07}} & {55.65{\tiny $\pm$ 0.07}} \\
		{\acastle} & 66.24{\tiny $\pm$0.15} &  78.33{\tiny $\pm$0.09} & {43.63{\tiny $\pm$ 0.08}} & {56.33{\tiny $\pm$ 0.06}}\\
		\bottomrule
	\end{tabular}
	\label{tab:incremental}
\end{table}

\subsection{Comparison with One-Phase Incremental Learning Methods}
The inductive generalized few-shot learning is also related to the one-phase incremental learning~\cite{Li2018Learning,Li2018Learning,Liu2020Mnemonics}, where a model is required to adapt itself to the open set environment. In other words, after training over the closed set categories, a classifier should be updated based on the data with novel distributions or categories accordingly. One important thread of incremental learning methods relies on the experience replay, where a set of the closed set instances is preserved and the classifier for all classes is optimized based on the saved and novel few-shot data. 
In our inductive GFSL, the \castle variants do not save \textsc{seen} class instances and rely on the neural dictionary to adapt the classifier for a joint classification. Thus, \castle variants have lower computational (time) costs during the inference stage.

Towards comprehensive comparisons, we also investigate two popular incremental learning methods, \ie., LwF~\cite{Li2018Learning} and iCARL\cite{Li2018Learning}. We randomly save 5 images per \textsc{seen} class for both methods. By combining the stored images and the newly given \textsc{unseen} class images together, the model will be updated based on a cross-entropy loss and a distillation loss~\cite{Hinton2015Distilling}. We tune the balance weight between the classification and distillation loss, the initial learning rate for fine-tuning, and the optimization steps for both methods over the validation set. The harmonic mean accuracy in various evaluation scenarios over 10,000 tasks is listed in Table~\ref{tab:incremental}.

In our empirical evaluations, we find that incremental learning methods can get better results than our baselines since it fine-tunes the model with the distillation loss. However, their results are not stable since there are many hyper-parameters. Compared with these approaches, our \castle variants still keep their superiority over all criteria.

\begin{table}[t]
	\tabcolsep 5pt
	\centering
	\caption{The light-weight model adaptation by fine-tuning the scale and bias weights based on the classifier initialization from \castle variants. The harmonic mean accuracy in different evaluation scenarios on {\it Mini}ImageNet are recorded. The superscript $\dagger$ denotes the method with another light-weight update step.}
	\begin{tabular}{@{\;}l@{\;}c@{\;}c@{\;}c@{\;}c@{\;}}
		\addlinespace
		\toprule
		{Classification on} & \multicolumn{ 2}{c}{5-Way} & \multicolumn{ 2}{c}{20-Way} \\
		{Setups} & 1-Shot & 5-Shot & 1-Shot & 5-Shot\\
		\midrule
		{\castle} & 66.22{\tiny $\pm$0.15} &  76.32{\tiny $\pm$0.09} & {43.06{\tiny $\pm$ 0.07}} & {55.65{\tiny $\pm$ 0.07}} \\
		{\castle}$^\dagger$ & 66.24{\tiny $\pm$0.15} &  76.43{\tiny $\pm$0.09} & {43.12{\tiny $\pm$ 0.07}} & {55.85{\tiny $\pm$ 0.07}} \\
		{\acastle} & 66.24{\tiny $\pm$0.15} &  78.33{\tiny $\pm$0.09} & {43.63{\tiny $\pm$ 0.08}} & {56.33{\tiny $\pm$ 0.06}}\\
		{\acastle}$^\dagger$ &  66.33{\tiny $\pm$0.15} &  78.93{\tiny $\pm$0.09} & {43.68{\tiny $\pm$ 0.08}} & {56.42{\tiny $\pm$ 0.06}}\\
		\bottomrule
	\end{tabular}
	\label{tab:adaptation}
\end{table}

\subsection{Light-Weight Adaptation on \castle Variants}
As shown in the previous subsection, directly fine-tuning the whole model is prone to over-fit even with another distillation loss. Inspired by~\cite{Sun2019Meta,Li2019Learning}, we consider a light-weight fine-tune step based on the synthesized classifier by \castle variants. In detail, we reformulate the model $\mathbf{W}^\top \vphi(\vx)$ as $\mathbf{W}^\top (({\bf 1} + {\rm scale})\cdot\vphi(\vx) + {\rm bias})$, where $\mathbf{W}$ is the classifier output by the neural dictionary, the ${\rm scale}\in\mathbb{R}^d$ and ${\rm bias}\in\mathbb{R}^d$ are additional learnable vectors, and ${\bf 1}$ is a size $d$ vector with all values equal 1.

Given a few-shot task with \textsc{unseen} class instances, the model will be updated in the following ways. 5 images per \textsc{seen} class are randomly selected, after freezing the backbone $\vphi$, the classifier $\mathbf{W}$, the scale, and the bias are optimized based on a cross-entropy loss over both stored \textsc{seen} and \textsc{unseen} classes images. We tune the initial learning rate and the optimization steps over the validation set.

The results of such model adaptation strategies are listed in Table~\ref{tab:adaptation}. With further model adaptation, both \castle and \acastle could be improved.

\begin{figure*}[tbp]
	\centering
	\small
	\begin{tabular}{cc}
		\begin{minipage}[h]{0.47\linewidth}
			\tabcolsep 2pt
			\includegraphics[width=1.0\textwidth]{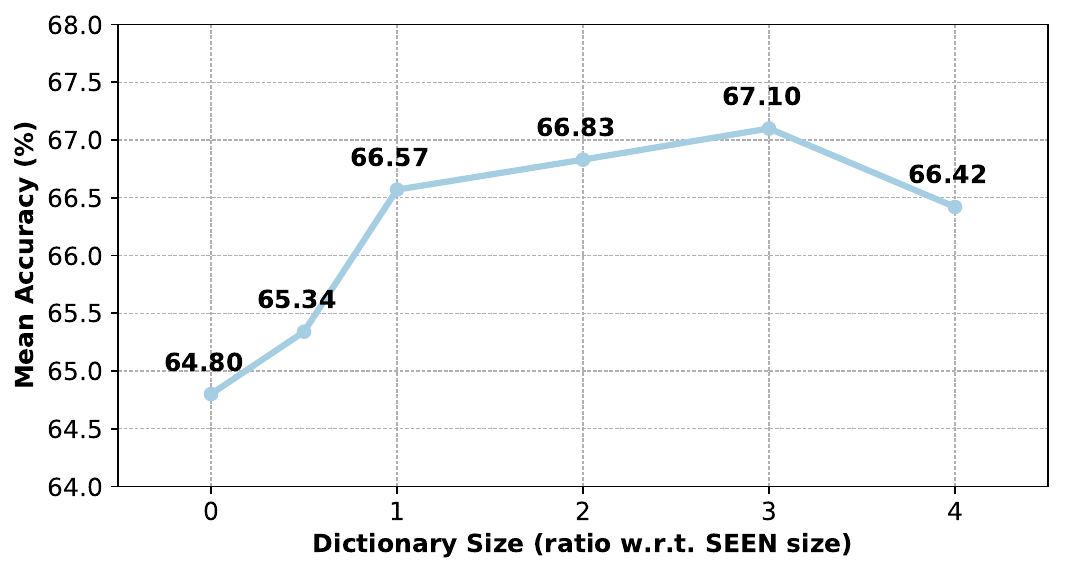}
			\caption{The 1-shot 5-way accuracy on \textsc{unseen} of {\it Mini}ImageNet with different size of dictionaries. }
			\label{fig:dict_size}
		\end{minipage}
		&
		\begin{minipage}[h]{0.47\linewidth}
			\tabcolsep 2pt
			\includegraphics[width=1.0\textwidth]{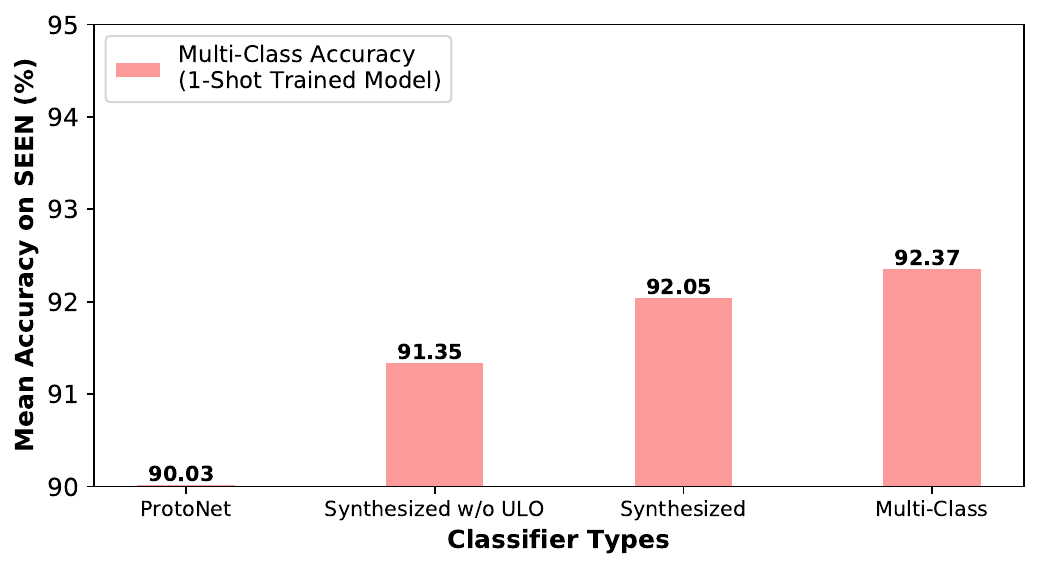}
			\caption{The 64-way multi-class accuracy on \textsc{seen} of {\it Mini}ImageNet with 1-shot trained model.}
			\label{fig:syncls}
		\end{minipage}
	\end{tabular}
\end{figure*}

\begin{table}[t]
	\tabcolsep 5pt
	\centering
	\caption{The performance with different choices of classifier synthesize strategies when tested with 5-Shot 5-Way \textsc{unsen} Tasks on {\it Mini}ImageNet. We denote the option compute embedding prototype and average synthesized classifiers as ``Pre-AVG'' and ``Post-AVG'' respectively.}
	\begin{tabular}{@{\;}l@{\;}c@{\;}c@{\;}}
		\addlinespace
		\toprule
		Perf. Measures  & FSL {\it Mean Acc.}  & GFSL {\it HM Acc.} \\
		\midrule
		{\castle} w/ Pre-AVG & 81.98 {\tiny $\pm$0.20} & 76.32 {\tiny $\pm$0.09} \\
		{\castle} w/ Post-AVG & 82.00 {\tiny $\pm$0.20} & 76.28 {\tiny $\pm$0.09}\\
		\bottomrule
	\end{tabular}
	\label{tab:post_avg}
\end{table}

\begin{table}[t]
	\tabcolsep 5pt
	\centering
	\caption{The GFSL performance (harmonic mean accuracy) change with different number of classifiers (\# of CLS) when tested with 1-Shot 5-Way \textsc{unsen} Tasks on {\it Mini}ImageNet. }
	\begin{tabular}{@{\;}l@{\;}c@{\;}c@{\;}c@{\;}c@{\;}}
		\addlinespace
		\toprule
		\# of Classifiers  & 1  & 64 & 128 & 256 \\
		\midrule
		{\castle} & 64.53 {\tiny $\pm$0.15} & 65.61 {\tiny $\pm$0.15} & 66.22 {\tiny $\pm$0.15}& 66.72 {\tiny $\pm$0.15}\\
		\bottomrule
	\end{tabular}
	\label{tab:multi_classifier}
\end{table}

\subsection{Effects on the neural dictionary size $|\calB|$} 
We show the effects of the dictionary size (as the ratio of \textsc{seen} class size 64) for the standard few-shot learning (measured by mean accuracy when there are 5 \textsc{unseen} classes) in Figure~\ref{fig:dict_size}. We observe that the neural dictionary with a ratio of 2 or 3 works best amongst all other dictionary sizes. Therefore, we fix the dictionary size as 128 across all experiments. Note that when $|\calB|=0$, our method degenerates to case optimizing the unified objective in Eq.~\ref{eq:gfsl_meta_obj} without using the neural dictionary (the \textsc{Castle}$^-$ model in \S~\ref{sec:exp_cross}). 

\subsection{How well is synthesized classifiers comparing with multi-class classifiers?} 
To assess the quality of the synthesized classifier, we made a comparison against ProtoNet and also the Multi-class Classifier on the head \textsc{seen} concepts. To do so, we sample few-shot training instances on each \textsc{seen} category to synthesize classifiers (or compute class prototypes for ProtoNet), and then use the synthesized classifiers/class prototypes solely to evaluate multi-class accuracy. The results are shown in Figure~\ref{fig:syncls}. We observe that the learned synthesized classifier outperforms over ProtoNet. Also, the model trained with the unified learning objective improves over the vanilla synthesized classifiers. Note that there is still a gap left against multi-class classifiers trained on the entire dataset. It suggests that the classifier synthesis we learned is effective against using sole instance embeddings.

\subsection{Different choices of the classifier synthesis}\label{sec:appendix-choice}
As in Eq.~\ref{eProto}, when there is more than one instance per class in a few-shot task (\ie., $K > 1$), {\castle} compute the averaged embeddings first, and then use the prototype of each class as the input of the neural dictionary to synthesize their corresponding classifiers. 
Here we explore another choice to deal with multiple instances in each class. We synthesize classifiers based on each instance first, and then average the corresponding synthesized classifiers for each class. This option equals an ensemble strategy to average the prediction results of each instance's synthesized classifier. We denote the pre-average strategy (the one used in {\castle}) as ``Pre-AVG'', and the post-average strategy as ``Post-AVG''. The 5-Shot 5-way classification results on {\it Mini}ImageNet for these two strategies are shown in Table~\ref{tab:post_avg}. From the results, ``Post-AVG'' does not improve the FSL and GFSL performance obviously. Since averaging the synthesized classifiers in a hindsight way costs more memory during meta-training, we choose the ``Pre-AVG'' option to synthesize classifiers when there are more than 1 shot in each class. In our experiments, the same conclusion also applies to {\acastle}.

\subsection{How is \textit{multiple classifiers learning's} impact over the training?}\label{sec:appendix-multi-classifier}
Both {\castle} and {\acastle} adopts a multi-classifier training strategy (as described in \S~\ref{sec:method}), \ie. considering multiple GFSL tasks with different combinations of classifiers in a single mini-batch. In Table~\ref{tab:multi_classifier}, we show the influence of the multi-classifier training method based on their GFSL performance (harmonic mean). It shows that with a large number of classifiers during the training, the performance of \castle asymptotically converges to its upper-bound. We find {\acastle} shares a similar trend. 

\begin{table}[t]
	\tabcolsep 5pt
	\centering
	\caption{The performance gap between \castle variants and a kind of ``many-shot'' upper bound (denoted as ``UB'') on {\it Mini}ImageNet. The ability of FSL classification is measured by the mean accuracy, while the harmonic mean accuracy is used as a criterion for GFSL. 5-Shot classification performance of \castle and \acastle are listed for a comparison.}
	\begin{tabular}{@{\;}l@{\;}c@{\;}c@{\;}c@{\;}c@{\;}}
		\addlinespace
		\toprule
		{Setups} & \multicolumn{ 2}{c}{5-Way} & \multicolumn{ 2}{c}{20-Way} \\
		{Measures} & FSL & GFSL &  FSL & GFSL \\
		\midrule
		{\castle} & 81.98 {\tiny $\pm$0.14} & 76.32 {\tiny $\pm$0.09} & 56.97 {\tiny $\pm$0.06} & 43.06 {\tiny $\pm$0.07} \\
		{\acastle} & 82.08 {\tiny $\pm$0.14} & 78.33 {\tiny $\pm$0.09} & 57.29 {\tiny $\pm$0.06} & 56.33 {\tiny $\pm$0.06} \\
		UB & 87.08 {\tiny $\pm$0.10} & 80.23 {\tiny $\pm$0.09} & 68.25 {\tiny $\pm$0.05} & 68.72 {\tiny $\pm$0.12}\\
		\bottomrule
	\end{tabular}
	\label{tab:upperbound}
\end{table}

\subsection{The gap to the performance ``Upper Bound'' (UB)}
We focus on the (generalized) few-shot learning scenario where there are only budgeted examples in the \textsc{unseen} class tasks. To show the potential improvement space in such tasks, we also investigate a kind of upper bound model where all the available images are used to build the \textsc{unseen} class classifier during the inference stage. 

We implement the upper bound model based on the ProtoNet, and the results are in Table~\ref{tab:upperbound}. Specifically, in the FSL classification scenario, all the \textsc{unseen} class images except those preserved for evaluation are used to build more precise prototypes, and the mean accuracy over 10,000 tasks are recorded; in the GFSL classification scenario, the many-shot \textsc{unseen} class images are utilized as well, and the calibrated harmonic mean is used as the performance measure.

Since the upper bound takes advantage of all the available training images for the few-shot categories, it performs better than the few-shot \castle and \acastle in all the scenarios. The gap between the few-shot learning methods and the upper bound becomes larger when more \textsc{unseen} classes (ways) are involved.

\bibliographystyle{spmpsci}      
\bibliography{references}
\end{document}